%% file: main.tex
\documentclass[10pt,twocolumn,letterpaper]{article}
\PassOptionsToPackage{table}{xcolor}
\usepackage{cvpr}
\usepackage{cuted}
\usepackage{currfile}
\usepackage{caption}

\usepackage[utf8]{inputenc}
\usepackage[T1]{fontenc}
\usepackage{amsfonts}
\usepackage{nicefrac}
\usepackage{microtype}
\usepackage{tabularx}
\usepackage{mathtools}
\usepackage{amsthm}
\usepackage{multirow}
\definecolor{lightgray}{gray}{0.9}
\definecolor{lightblue}{RGB}{220,235,255}
\AtBeginDocument{\hfuzz=0pt}
\usepackage{algorithm}
\usepackage{algpseudocode}

\usepackage{placeins}
\definecolor{cvprblue}{rgb}{0.21,0.49,0.74}
\usepackage[pagebackref,breaklinks,colorlinks,allcolors=cvprblue,hypertexnames=false]{hyperref}
\hypersetup{pdftitle={OnCache: Learning Dynamic Caching Schedules for Efficient Diffusion Inference},pdfauthor={Zhikang Xie, Xichen Ye, Yifan Wu, Yuxuan Liu, Qianlong Wang, Haoshen Yu, Peizhu Gong, Weizhong Zhang, Cheng Jin}}
\def\confName{CVPR}
\def\confYear{2027}

\title{OnCache: Learning Dynamic Caching Schedules for Efficient~Diffusion~Inference}

\author{%
  Zhikang Xie$^{1}$\thanks{Equal contribution.}\quad
  Xichen Ye$^{1}$\footnotemark[1]\quad
  Yifan Wu$^{1}$\quad
  Yuxuan Liu$^{1}$\quad
  Qianlong Wang$^{2}$\\
  Haoshen Yu$^{1}$\quad
  Peizhu Gong$^{1}$\quad
  Weizhong Zhang$^{2}$\quad
  Cheng Jin$^{1,3}$\thanks{Corresponding author.}\\
  {\normalsize $^{1}$College of Computer Science and Artificial Intelligence, Fudan University}\\
  {\normalsize $^{2}$School of Data Science, Fudan University\quad
  $^{3}$Shanghai Key Laboratory of Intelligent Information Processing}
}

\begin{document}
\maketitle
\setlength{\stripsep}{0pt}
\begin{strip}
\centering
\includegraphics[width=0.80\textwidth]{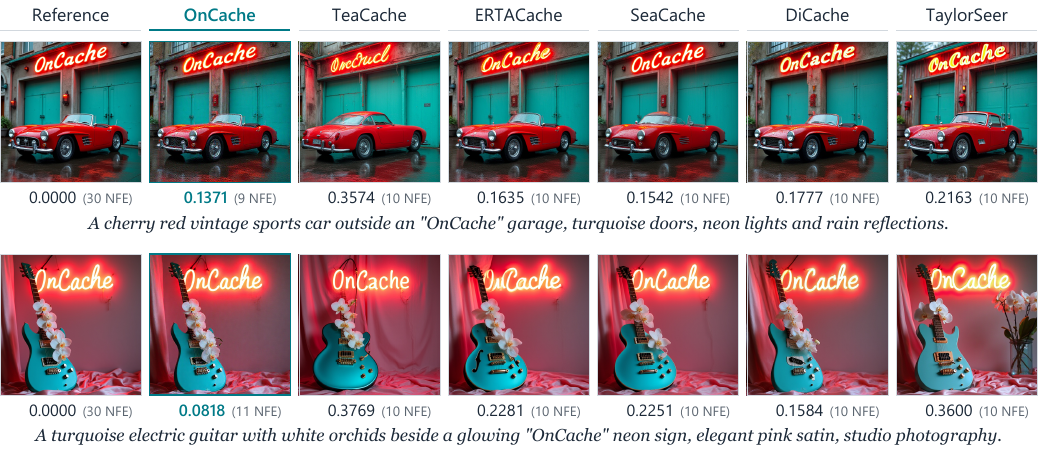}
\captionof{figure}{\textbf{Preserving structure and detail with OnCache.} Values give LPIPS to Reference ($\downarrow$), with actual computation in parentheses.}
\label{fig:teaser}
\par\vspace{4pt}
\end{strip}
\patchcmd{\abstract}{\vspace*{12pt}\noindent}
  {\vspace*{6pt}\nointerlineskip\noindent}{}
  {\PackageError{teaser}{Could not adjust Abstract heading spacing}{}}
\begin{abstract}
Diffusion models have revolutionized generative tasks but incur high inference latency due to iterative denoising. Cache-based strategies accelerate inference by reusing intermediate features, yet they largely rely on static, sample-agnostic schedules. Our empirical analysis reveals two factors that these schedules overlook. (i) denoising dynamics and computational requirements vary across prompts, motivating adaptive computation allocation. (ii) sensitivity to caching errors varies across timesteps, so static schedules may reuse features at error-sensitive steps or waste computation at less sensitive ones. 
We therefore propose \textbf{OnCache}, a dynamic caching framework that makes prompt-adaptive decisions online about how much computation to allocate and when to refresh cached features. A \textbf{policy} is trained offline on paired compute and cache continuations from identical denoising states, using differences in terminal fidelity and the number of function evaluations (NFE) to account for the downstream consequences of each decision. To further improve the generation quality, a neural \textbf{corrector} compensates for caching-induced errors and mitigates their accumulation throughout the denoising trajectory. Both components are lightweight and introduce negligible additional inference latency. 
On \textsc{FLUX.1-dev}, OnCache achieves a nearly $3\times$ generation speedup at $1024^2$ resolution, with LPIPS of $0.156$ and SSIM of $0.860$. We further extend OnCache to video generation on \textsc{Wan2.1-T2V-1.3B}, achieving state-of-the-art efficiency--quality trade-offs across both modalities.
\end{abstract}

\section{Introduction}
\label{sec:intro}
\begin{figure*}[t]
    \centering
    \includegraphics[width=0.90\linewidth]{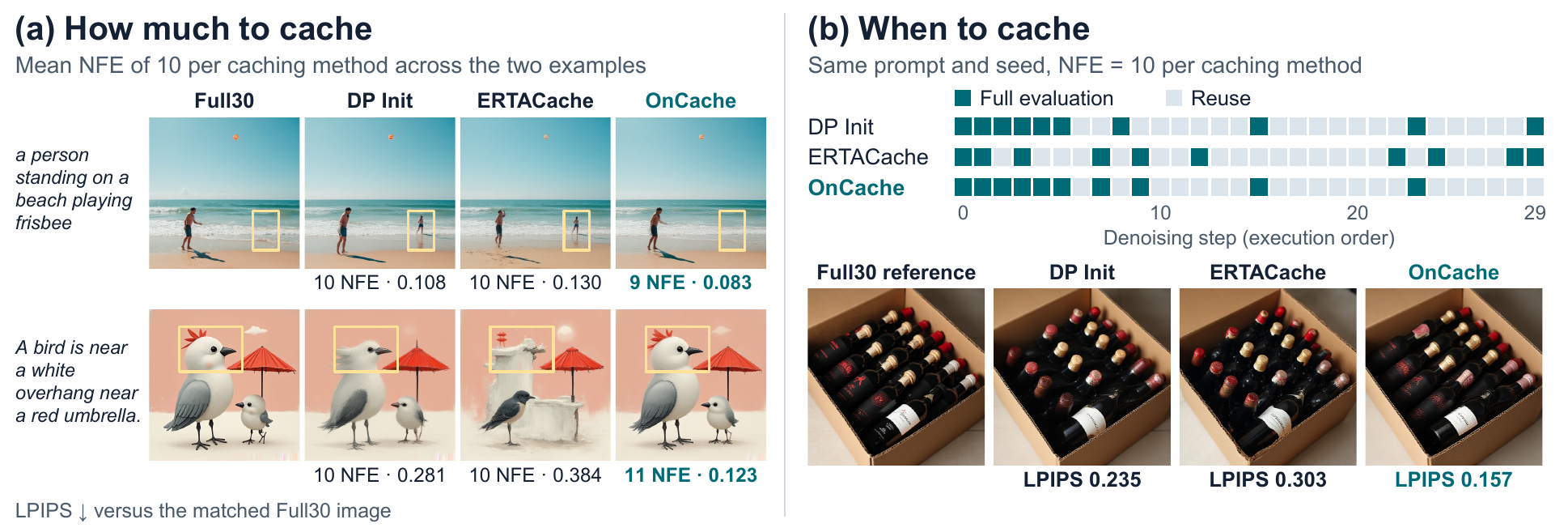}
    \caption{\textbf{A computation budget can be reallocated across samples and timesteps.}
    \textbf{(a)} Moving one full evaluation between two samples improves
    both outputs at unchanged aggregate computation.
    \textbf{(b)} Changing refresh locations improves final-image fidelity for the same prompt (with the same total NFE). Lower LPIPS indicates closer agreement with the full reference.}
    \label{fig:motivation}
\end{figure*}

Diffusion models have become a leading paradigm for visual generation, producing high-quality images~\citep{DBLP:conf/nips/HoJA20,rombach2022latent} and videos~\citep{DBLP:conf/iclr/YangTZ00XYHZFYZ25}. Modern systems increasingly combine iterative denoising with scalable Transformer backbones (DiTs)~\citep{DBLP:conf/iccv/PeeblesX23,fluxdevmodelcard}, requiring repeated passes through a large network for each generated output. As model capacity and generation quality grow, this repeated computation becomes a substantial inference bottleneck. Reducing its cost while retaining the quality of full-step generation is therefore central to the practical deployment of diffusion models.

Diffusion acceleration addresses this cost from several directions. Improved samplers reduce denoising evaluations through implicit updates and higher-order numerical integration~\citep{song2021ddim,DBLP:conf/iclr/Liu0LZ22,DBLP:conf/nips/KarrasAAL22,lu2022dpmsolver,DBLP:journals/ijautcomp/LuZBCLZ25,DBLP:conf/nips/ZhaoBR0L23}. Distillation compresses pretrained diffusion models into few-step or one-step generators through progressive, consistency-based, and distribution-matching objectives~\citep{salimans2022progressive,DBLP:conf/icml/SongD0S23,DBLP:journals/corr/abs-2310-04378,DBLP:conf/iclr/LuS25,DBLP:conf/cvpr/YinG0SDFP24,DBLP:conf/nips/YinGP0SDF24}. Feature caching offers a complementary approach by exploiting temporal redundancy within the denoiser~\citep{DBLP:conf/cvpr/MaFW24,DBLP:conf/cvpr/WimbauerWSDHHSZ24}. Intermediate representations can be reused across neighboring timesteps, bypassing expensive computation while keeping the pretrained backbone fixed. The resulting efficiency--quality trade-off depends on the amount of cache-reuse and the placement of full evaluations that refresh the cache.

Many feature caching methods determine a shared cache-refresh schedule offline. ERTACache identifies reusable timesteps through residual error profiling, while DPCache and BudCache optimize calibrated path costs and final-output quality, respectively~\citep{peng2026ertacache,cui2026dpcache,lei2026budcache}. More adaptive methods, including TeaCache, SeaCache, and DiCache, adjust refresh decisions along the running trajectory using input changes, spectral information, or shallow probes~\citep{liu2025teacache,chung2026seacache,bu2026dicache}. BAG further learns online gates from searched prompt--budget schedules while enforcing a prescribed per-sample computation~\citep{DBLP:journals/corr/abs-2608-09231}. These approaches make refresh timing responsive to individual trajectories, but adaptation remains driven by local proxy signals or constrained to allocating a fixed budget within each sample. A more adaptive online policy must jointly determine how much computation each sample receives and when to spend it, according to the downstream fidelity benefit of each refresh. We identify two forms of heterogeneity that motivate this joint allocation across samples and timesteps, as illustrated in Figure~\ref{fig:motivation}:

\begin{itemize}
    \item \textbf{Sample-level heterogeneity.} Different prompts and denoising trajectories tolerate different amounts of reuse. In Fig.~\ref{fig:motivation}a, \hyperref[sec:dp-init]{DP Init} (OnCache's initial static schedule, optimized offline via dynamic programming) and ERTACache each use ten NFE for both prompts. OnCache uses nine for the beach scene and eleven for the bird scene, producing outputs closer to their full-step references with the same total twenty NFE. The useful computation budget therefore varies across samples: allocating a larger NFE where it improve fidelity and reusing features more aggressively elsewhere can improve quality under the same aggregate NFE budget. This is the question of \emph{how much to cache}.
    \item \textbf{Timestep-level heterogeneity.} Feature reuse has different consequences at different points in a denoising trajectory. Fig.~\ref{fig:motivation}b holds the prompt, initial noise, and ten-NFE budget constant. By selecting different refresh locations, OnCache reduces final-image LPIPS to 0.157, compared with 0.235 for DP Init and 0.303 for ERTACache. Thus, under a fixed NFE budget, refreshes should be spent at states where fresh features most improve final-image fidelity by reducing the downstream impact of reuse errors. This is the question of \emph{when to cache}.
\end{itemize}

Together, these observations motivate learning \emph{how much to cache} across samples and \emph{when to cache} within each trajectory. An average computation budget allows full evaluations to move between samples as well as between timesteps. The two decisions are coupled through the denoising dynamics: refreshing the cache changes the current prediction, subsequent latent states, and later caching decisions. A refresh costs one full evaluation now but can save evaluations later by enabling more reuse. An effective policy must therefore anticipate how the current choice between compute and cache affects both final-image fidelity and the total NFE.

We propose \textbf{OnCache}, a dynamic caching framework that allocates computation \textbf{on}line across samples and timesteps according to the downstream value of each refresh. Trained offline with the diffusion backbone frozen, a compact policy observes the current denoising state and cache history to choose between full computation and cache-reuse at runtime. Its supervision comes from \emph{paired continuations}: starting from an identical state, we execute the two alternative actions and let the same policy complete each branch on its resulting states. The difference in terminal fidelity to a full-step reference measures the benefit of refreshing, while the difference in total NFE measures its cost. Renewed paired queries refine the policy using the states it visits as its behavior changes. A global offset sets the average budget, allowing both NFE and refresh locations to vary across samples.

At cache steps, errors between cache and compute features propagate through later denoising updates. OnCache pairs the policy with a lightweight neural corrector that uses residual history and current input drift to compensate for errors in reused features. With the policy fixed, we query the full backbone at selected states along its denoising trajectories to obtain target features and predictions. These targets provide supervision for training the corrector. At inference, the corrector adjusts cached features, and the policy adapts its caching decisions online to the resulting trajectory. Together, our proposed OnCache improves the efficiency--quality trade-off by adapting computation online and reducing local errors from feature reuse.

Our core contributions can be summarized as follows:
\begin{itemize}
    \item We formulate diffusion caching as sequential online allocation under an average budget, enabling a learned policy to adapt both how much to cache across samples and when to cache along each denoising trajectory.
    \item We develop paired-continuation supervision that connects individual decisions to terminal fidelity gains and NFE costs, together with a separately trained  corrector that further improves final generation quality.
    \item Experiments on image and video generation demonstrate state-of-the-art efficiency--quality trade-offs compared with strong recent baselines, validating the effectiveness of dynamic online computation allocation.
\end{itemize}
\section{Related Work}
\label{sec:related}
\paragraph{Cache reuse and correction.}
Feature caching reuses intermediate representations across denoising
steps~\citep{DBLP:conf/cvpr/MaFW24,DBLP:conf/cvpr/WimbauerWSDHHSZ24}.
TaylorSeer and Spectrum predict feature evolution from cached
history~\citep{liu2025taylorseer,han2026spectrum}, while ERTACache combines residual rectification with timestep adjustment~\citep{peng2026ertacache}.

\vspace{-12pt}
\paragraph{Adaptive caching.}
TeaCache, SeaCache, and DiCache adapt refresh decisions using input
changes, spectral evolution, and shallow probes,
respectively~\citep{liu2025teacache,chung2026seacache,bu2026dicache}.
EpaCache calibrates timestep tolerances from the terminal impact of
isolated cache actions~\citep{DBLP:journals/corr/abs-2608-29264}.
OnCache learns final-image fidelity gains and NFE differences from paired
continuations that follow the same policy after the initial action.

\vspace{-12pt}
\paragraph{Learned and budgeted scheduling.}
DPCache optimizes calibrated path costs, BudCache searches shared schedules, and ReCache learns them from terminal rewards~\citep{cui2026dpcache,lei2026budcache,DBLP:journals/corr/abs-2606-06060}. BAG distills searched prompt--budget schedules into online gates with exact per-image NFE~\citep{DBLP:journals/corr/abs-2608-09231}. Our proposed OnCache initializes data collection with DP and learns state-dependent decisions under an average budget, adapting both NFE and refresh timing across samples.  Appendix~\ref{app:related} covers broader acceleration
methods and detailed comparisons.
\section{Methodology}
\label{sec:method}
\begin{figure*}[t]
    \centering
    \includegraphics[width=0.86\linewidth,height=0.376\textwidth]{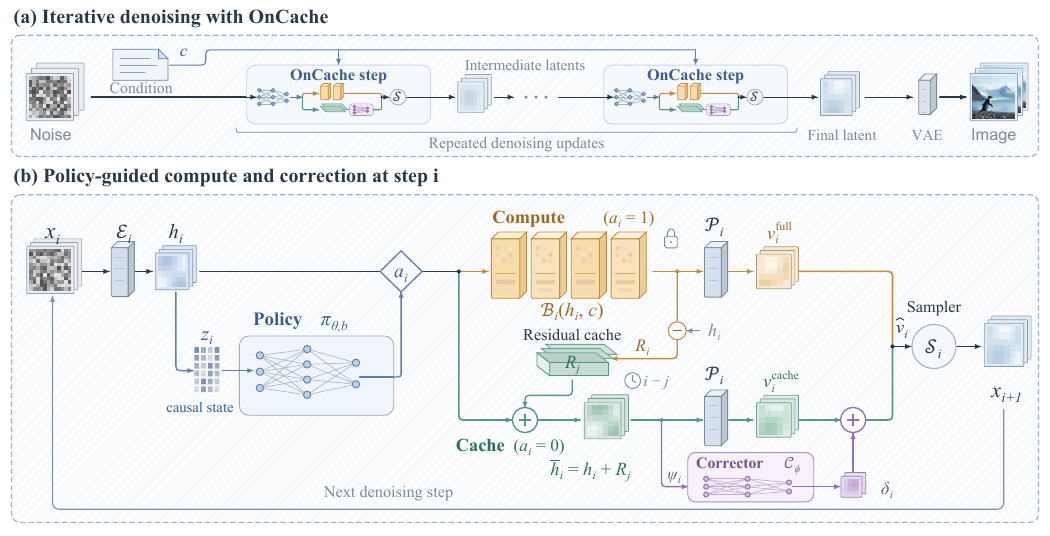}
    \caption{\textbf{OnCache inference framework.} (a) Conditioned generation repeatedly applies an OnCache denoising step, producing intermediate latents. (b) The policy $\pi_{\theta,b}$ observes the causal state $z_i$ and selects full computation ($a_i=1$) or cache reuse ($a_i=0$). Full computation refreshes the residual cache with $R_i=\mathcal B_i(h_i,c)-h_i$; cache reuse forms $\bar h_i=h_i+R_j$, where $j$ is the most recent refresh index. The corrector adjusts reused features before the output head; $\delta_i$ denotes the resulting adjustment to the denoiser output.}
    \label{fig:method}
    \vspace{-3pt}
\end{figure*}

OnCache pairs an adaptive policy that allocates full evaluations under an average budget with a corrector that adjusts reused features. We first obtain a static schedule with dynamic programming, then optimize the policy from paired continuations, and finally train the corrector with the policy fixed. All training is offline with the diffusion
backbone frozen. Figure~\ref{fig:method} shows the inference procedure.

\subsection{Problem Formulation}
\label{sec:problem-formulation}
\begingroup
\setlength{\abovedisplayskip}{3pt}
\setlength{\belowdisplayskip}{3pt}
\setlength{\abovedisplayshortskip}{3pt}
\setlength{\belowdisplayshortskip}{3pt}
We consider a diffusion-based generative process that progressively
denoises a latent variable from $x_0$ to $x_N$ over $N$ sampling steps.
At step $i\in\{0,\ldots,N-1\}$, the denoiser takes the current noisy
sample $x_i$ and conditioning $c$. We write it as an input embedding
$\mathcal E_i$, a Transformer stack $\mathcal B_i$, and an output head
$\mathcal P_i$. Given $h_i=\mathcal E_i(x_i,c)$, the Transformer output
can be expressed in residual form as:
\begin{equation}
\mathcal B_i(h_i,c)=h_i+R_i,\qquad
R_i=\mathcal B_i(h_i,c)-h_i.
\end{equation}
The denoiser prediction is $v_i=\mathcal P_i(h_i+R_i)$, and the sampler
advances to $x_{i+1}=\mathcal S_i(x_i,v_i)$. Previous work, such as
ERTACache~\citep{peng2026ertacache}, shows that residual features often evolve smoothly across adjacent timesteps. This temporal redundancy suggests approximating $R_i$ with the cached residual $R_j$ from the most recent full evaluation at step $j<i$.

We introduce a binary action $a_i\in\{0,1\}$ at each step: $a_i=1$ performs full backbone computation, and $a_i=0$ reuses the cached residual. The update rule is:
\begin{equation}
\bar h_i=h_i+
\begin{cases}
 R_i,&a_i=1,\\
 R_j,&a_i=0,
\end{cases}
\qquad v_i=\mathcal P_i(\bar h_i).
\end{equation}
When $a_i=1$, the exact residual $R_i$ refreshes the cache. When $a_i=0$, reusing $R_j$ bypasses the Transformer stack. With
$\mathrm{NFE}=\sum_{i=0}^{N-1}a_i$ counting full evaluations, OnCache selects actions to preserve final-output fidelity under an average computation budget across samples.

\endgroup
\subsection{Policy for Adaptive Computation}
\label{sec:cache}

\paragraph{Policy Learning Motivation.}
Under an average NFE budget, the policy must decide whether a refresh
is worth its cost. A refresh changes the current prediction and later
policy decisions: computing now may enable more reuse later. Its benefit
is the change in final fidelity, while its cost is the change in total NFE,
not necessarily one extra evaluation. In the final fitting bank, refreshing
leads to the same or fewer total NFE than reuse in 797 of 1,792 paired
continuations (44.5\%; 95\% prompt-bootstrap interval 41.9--47.0\%;
Fig.~\ref{fig:method-diagnostics}a). We therefore learn the final-fidelity
gain and total NFE change from paired compute and cache continuations.

\par\noindent
\begin{minipage}{\columnwidth}
    \centering
    \includegraphics[width=0.90\linewidth,trim=1 7 0 1,clip]{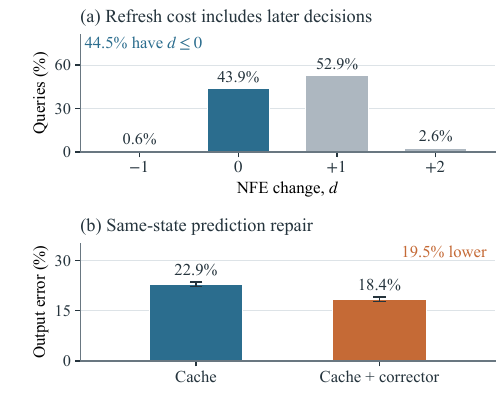}
    \vspace{-2pt}\captionof{figure}{\textbf{Continuation costs and same-state correction.}
    (a) NFE changes from 1,792 queries on 256 prompts at native $256^2/512^2$ resolution. (b) Mean relative output error
    $\|v-v^\star\|_2/\|v^\star\|_2$ on 640 same-state queries from 80 corrector-validation trajectories. Error bars give 95\% prompt-cluster bootstrap intervals (5,000 resamples).}
    \label{fig:method-diagnostics}
\end{minipage}\par
\vspace{-5pt}

\paragraph{Dynamic programming initialization.}
\label{sec:dp-init}
We initialize policy learning with a shared schedule of $K$ full evaluations
over $N$ denoising steps. Following prior work on DP-based cache
scheduling~\citep{cui2026dpcache}, we use dynamic programming (DP) to find a schedule that minimizes a residual-mismatch proxy. This gives policy learning an informed starting point rather than an arbitrary refresh schedule. On full-compute trajectories for profiling prompts $p\in\mathcal D_{\mathrm{DP}}$,
we form residual summaries $\Xi_{p,t}$ by subsampling tokens and channels. Reusing the residual from step $i$ at step $t$ incurs the proxy error:
\begin{equation}
 e(i,t)=
 \left(\frac{t+1}{N}\right)^\gamma
 \frac{1}{|\mathcal D_{\mathrm{DP}}|}
 \sum_{p\in\mathcal D_{\mathrm{DP}}}
 \frac{\operatorname{MS}(\Xi_{p,t}-\Xi_{p,i})}
 {\operatorname{MS}(\Xi_{p,t})+\epsilon}.
\label{eq:dp-mismatch}
\end{equation}
Here $\operatorname{MS}$ is the mean squared entry, $\gamma$ controls time weighting, and $\epsilon$ stabilizes normalization. The interval cost $C(i,j)=\sum_{t=i+1}^{j-1}e(i,t)$ sums errors over skipped steps.

Let $D[k,j]$ be the minimum proxy cost of reaching refresh $j$ with $k$ full evaluations. The initial table $D_0$ stores the prescribed boundary costs and is infinite elsewhere. For admissible predecessors $\mathcal I_j$, the recurrence is:
\begin{equation}
 D[k,j]=\min\!\left\{D_0[k,j],\,\min_{i\in\mathcal I_j}
 \bigl(D[k-1,i]+C(i,j)\bigr)\right\}.
\label{eq:dp-recursion}
\end{equation}
Backtracking recovers the schedule. We follow this schedule
to collect initial trajectory states, then construct paired compute/cache
continuations from those states to train the policy. The schedule also serves
as a static control. Appendix~\ref{app:dp-init} gives the profiling, search and pseudo code details.

\paragraph{Policy learning.}
At each cache-eligible step, the policy observes the current input $h_i$, conditioning $c$, sampling progress $i$, and trajectory history $\mathcal H_i$, which contains past inputs, actions, and cached residuals.
It constructs a causal state $z_i$ and dynamically selects a binary action:
\begin{equation}
\begin{aligned}
 z_i&=\varphi(h_i,c,i,\mathcal H_i),\\
 a_i&=\pi_{\theta,b}(z_i)=\mathbf1_{\{s_\theta(z_i)+b\geq0\}}.
\end{aligned}
\label{eq:score}
\end{equation}
Here $\varphi$ is a fixed state-construction map that summarizes the inputs using pooling, statistics, and projections, capturing input drift and residual changes. It uses information available before the current backbone evaluation, and only full evaluations update the cached residual history. The score $s_\theta$ determines whether to refresh, while the shared offset $b$ controls the average computation budget.
We encode the DP schedule as a timestep prior that raises the policy score at its refresh steps and lowers it at other eligible steps, guiding the initial policy toward this schedule.

To learn the score $s_\theta$, we collect states along the DP schedule initially and along the current policy in later rounds. At each selected state, we start paired continuations with opposite actions: one refreshes and the other reuses the cache. Both then follow the same frozen collection strategy on
their respective states until generation completes. For action $a$, let
$\mathbf m_i^{(a)}$ be the resulting output's scaled fidelity vector relative
to the full-step reference, comprising negated LPIPS~\citep{DBLP:conf/cvpr/ZhangIESW18},
SSIM~\citep{DBLP:journals/tip/WangBSS04}, and PSNR so that higher values indicate better fidelity. Let $\mathrm{NFE}_i^{(a)}$ denote the total number of full evaluations in that branch. The paired targets are the
fidelity gain and change in total computation:
\begin{equation}
\begin{aligned}
 \mathbf g_i&=\mathbf m_i^{(1)}-\mathbf m_i^{(0)},\\
 d_i&=\mathrm{NFE}_i^{(1)}-\mathrm{NFE}_i^{(0)}.
\end{aligned}
\label{eq:targets}
\end{equation}
The shared prefix cancels in $d_i$, leaving the effect of the current action
and all subsequent decisions. A refresh can therefore have zero or negative
net NFE cost if it leads to sufficiently fewer later full evaluations.

Let $\mathcal Q$ be the set of collected paired queries. Each query
$q\in\mathcal Q$ provides a state $z_q$, fidelity gain $\mathbf g_q$, and
NFE change $d_q$. A linear predictor $f_W$ estimates the concatenation of
these targets from state--time features. We fit its coefficients $W$ by
weighted, regularized least squares:
\begin{equation}
 W^*=\arg\min_W\Bigl[\sum_{q\in\mathcal Q}u_q
 \bigl\|f_W(z_q)-[\mathbf g_q^\top,d_q]^\top\bigr\|_2^2
 +\|W\|_\Lambda^2\Bigr].
\label{eq:value-regression}
\end{equation}
Here $u_q$ weights the contribution of query $q$. The regularization term
$\|W\|_\Lambda^2$ is a weighted sum of squared coefficients, with the
diagonal matrix $\Lambda$ setting the penalty strength for each feature.
For state $z_i$, the fitted predictor estimates the fidelity gain
$\hat{\mathbf g}_i$ and NFE change $\hat d_i$. These predictions and the
DP time prior form the initial policy score.

As the policy visits new states, we collect new paired continuations and refit the predictor. We combine its predicted fidelity gain and NFE change into a net refresh value:
\begin{equation}
 V_i=\mathbf w^\top\hat{\mathbf g}_i-\lambda\hat d_i.
\label{eq:policy-value}
\end{equation}
Here $\mathbf w$ weights the predicted fidelity gains and $\lambda$ controls
the contribution of the predicted NFE change. A positive $V_i$ favors
refreshing over reuse under this fidelity--computation trade-off; a negative
$V_i$ favors reuse.

With $b$ fixed, let $s_{\rm old}$ and $s_{\rm new}$ denote the policy scores
before and after the update. Their decision scores satisfy:
\begin{equation}
\begin{aligned}
 \underbrace{s_{\rm new}(z_i)+b}_{\text{new score}}
 ={}&\underbrace{s_{\rm old}(z_i)+b}_{\text{old score}}\\
 &+\alpha\,\operatorname{clip}\!\left(
 \underbrace{V_i}_{\text{net value}}
 -\underbrace{(s_{\rm old}(z_i)+b)}_{\text{old score}}\right).
\end{aligned}
\label{eq:decision-score-update}
\end{equation}
The difference inside the clip compares the predicted net value with the
old decision score, shifting the score in the direction of the predicted value.
The strength $0<\alpha \leq1$ controls the update, while the clip limits its
magnitude. For example, when $V_i>0$ but the old decision score is negative,
the update raises the score and can change the action from reuse to refresh
in Eq.~\ref{eq:score}. 
Subtracting $b$ gives the equivalent update of the policy score itself:
\begin{equation}
 s_\theta(z_i)\leftarrow{}s_\theta(z_i)+\alpha\,\operatorname{clip}\!\bigl(V_i-s_\theta(z_i)-b\bigr).
\label{eq:policy-update}
\end{equation}
For inference, the policy and its shared offset $b$ are frozen, while decisions remain state-dependent
and NFE can vary across samples. Appendix~\ref{app:value-fit} gives further
details.

\begin{table*}[t]
\centering
\footnotesize
\setlength{\tabcolsep}{2pt}
\renewcommand{\arraystretch}{1.05}
\caption{FLUX.1-dev on 1,000 COCO captions at 30 sampling steps. Each cell reports $512^2$ / $1024^2$ resolution results. OnCache (w/o) disables the corrector; OnCache (w) is the full method. Bold/underlined scores mark the best/second-best results at each resolution.}
\label{tab:flux-cache-comparison}
\vspace{-5pt}
\begin{tabularx}{\textwidth}{@{}>{\raggedright\arraybackslash}p{0.10\textwidth}*{7}{>{\centering\arraybackslash}X}@{}}
\toprule
\multirow{2}{*}{Method} & \multicolumn{3}{c}{\textbf{Efficiency}} & \multicolumn{4}{c}{\textbf{Performance}} \\
\cmidrule(lr){2-4}\cmidrule(lr){5-8}
 & \mbox{Mean NFE} & \mbox{Gen. (s) $\downarrow$} & \mbox{Speedup $\uparrow$} & \mbox{LPIPS $\downarrow$} & \mbox{SSIM $\uparrow$} & \mbox{PSNR $\uparrow$} & \mbox{ImageReward$\uparrow$} \\
\midrule
Full30 & \mbox{30.000\,/\,30.000} & \mbox{6.064\,/\,19.455} & \mbox{1.00$\times$\,/\,1.00$\times$} & \mbox{--\,/\,--} & \mbox{--\,/\,--} & \mbox{--\,/\,--} & \mbox{0.965\,/\,1.040} \\
TeaCache & \mbox{10.033\,/\,10.040} & \mbox{2.118\,/\,6.701} & \mbox{2.86$\times$\,/\,2.90$\times$} & \mbox{0.349\,/\,0.397} & \mbox{0.663\,/\,0.685} & \mbox{17.48\,/\,17.00} & \mbox{0.929\,/\,0.961} \\
TaylorSeer & \mbox{10.000\,/\,10.000} & \mbox{2.370\,/\,7.269} & \mbox{2.56$\times$\,/\,2.68$\times$} & \mbox{0.219\,/\,0.232} & \mbox{0.753\,/\,0.776} & \mbox{20.21\,/\,20.34} & \mbox{0.937\,/\,0.985} \\
ERTACache & \mbox{10.000\,/\,10.000} & \mbox{2.097\,/\,6.659} & \mbox{2.89$\times$\,/\,2.92$\times$} & \mbox{0.214\,/\,0.266} & \mbox{0.768\,/\,0.771} & \mbox{21.21\,/\,20.82} & \mbox{0.956\,/\,1.000} \\
SeaCache & \mbox{10.003\,/\,9.982} & \mbox{2.127\,/\,6.689} & \mbox{2.85$\times$\,/\,2.91$\times$} & \mbox{0.179\,/\,0.217} & \mbox{0.805\,/\,0.810} & \mbox{22.64\,/\,22.17} & \mbox{0.951\,/\,0.991} \\
DiCache & \mbox{10.006\,/\,10.029} & \mbox{2.182\,/\,6.922} & \mbox{2.78$\times$\,/\,2.81$\times$} & \mbox{0.186\,/\,0.206} & \mbox{0.800\,/\,0.823} & \mbox{22.95\,/\,23.39} & \mbox{0.941\,/\,0.978} \\
DPCache & \mbox{10.000\,/\,10.000} & \mbox{2.238\,/\,6.933} & \mbox{2.71$\times$\,/\,2.81$\times$} & \mbox{0.163\,/\,0.198} & \mbox{0.805\,/\,0.804} & \mbox{22.54\,/\,22.17} & \mbox{0.958\,/\,0.989} \\
Spectrum & \mbox{10.000\,/\,10.000} & \mbox{2.117\,/\,6.725} & \mbox{2.86$\times$\,/\,2.89$\times$} & \mbox{0.147\,/\,0.215} & \mbox{0.819\,/\,0.803} & \mbox{24.23\,/\,23.83} & \mbox{\underline{0.983}\,/\,\underline{1.003}} \\
BudCache & \mbox{10.000\,/\,10.000} & \mbox{2.095\,/\,6.657} & \mbox{2.89$\times$\,/\,2.92$\times$} & \mbox{\underline{0.143}\,/\,0.185} & \mbox{0.828\,/\,0.832} & \mbox{24.77\,/\,\underline{24.78}} & \mbox{0.968\,/\,0.992} \\
\midrule
DP Init & \mbox{10.000\,/\,10.000} & \mbox{2.091\,/\,6.668} & \mbox{2.90$\times$\,/\,2.92$\times$} & \mbox{0.158\,/\,0.201} & \mbox{0.824\,/\,0.827} & \mbox{24.35\,/\,24.29} & \mbox{0.929\,/\,0.974} \\
\rowcolor{lightblue}
OnCache (w/o) & \mbox{9.867\,/\,10.175} & \mbox{2.098\,/\,6.771} & \mbox{2.89$\times$\,/\,2.87$\times$} & \mbox{0.144\,/\,\underline{0.176}} & \mbox{\underline{0.835}\,/\,\underline{0.841}} & \mbox{\underline{24.92}\,/\,24.69} & \mbox{0.950\,/\,0.994} \\
\rowcolor{lightblue}
OnCache (w) & \mbox{9.856\,/\,10.164} & \mbox{2.113\,/\,6.813} & \mbox{2.87$\times$\,/\,2.86$\times$} & \mbox{\textbf{0.127}\,/\,\textbf{0.156}} & \mbox{\textbf{0.857}\,/\,\textbf{0.860}} & \mbox{\textbf{25.55}\,/\,\textbf{25.52}} & \mbox{\textbf{0.989}\,/\,\textbf{1.007}} \\
\bottomrule
\end{tabularx}
\end{table*}

\begin{table*}[t]
\centering
\begin{minipage}{\textwidth}
\centering
\footnotesize
\setlength{\tabcolsep}{2pt}
\renewcommand{\arraystretch}{1.02}
\caption{Wan2.1-T2V-1.3B with 50 steps, $832\times480$ resolution, 81 frames, and five seeds. Quality scores for TeaCache and ERTACache are taken from the ERTACache paper~\citep{peng2026ertacache}. We remeasured their runtimes in our environment using 15 prompts. The 95\% bootstrap confidence intervals are 173.26--173.60\,s and 173.77--174.25\,s, respectively.}
\label{tab:wan-cache-comparison}
\vspace{-5pt}
\begin{tabularx}{\linewidth}{@{}>{\raggedright\arraybackslash}p{0.15\linewidth}*{7}{>{\centering\arraybackslash}X}@{}}
\toprule
\multirow{2}{*}{Method} & \multicolumn{3}{c}{\textbf{Efficiency}} & \multicolumn{4}{c}{\textbf{Performance}} \\
\cmidrule(lr){2-4}\cmidrule(lr){5-8}
 & \mbox{Mean NFE} & \mbox{Gen. (s) $\downarrow$} & \mbox{Speedup $\uparrow$} & \mbox{VBench $\uparrow$} & \mbox{LPIPS $\downarrow$} & \mbox{SSIM $\uparrow$} & \mbox{PSNR $\uparrow$} \\
\midrule
Full50 & 50.000 & 337.575 & 1.00$\times$ & 81.44\% & -- & -- & -- \\
TeaCache & 23.741 & 173.428 & 1.95$\times$ & 76.04\% & 0.291 & 0.569 & 16.17 \\
ERTACache & 24.000 & 174.017 & 1.94$\times$& 80.73\% & 0.110 & 0.820 & 23.77 \\
TaylorSeer & 20.000 & 151.119& 2.23$\times$ & 80.55\% & 0.280 & 0.606 & 16.68 \\
SeaCache & 20.007 & 148.565 & 2.27$\times$ & 81.04\% & 0.138 & 0.784 & 22.26 \\
DiCache & 23.718 & 193.889 & 1.74$\times$ & \underline{81.10\%} & \underline{0.103} & \underline{0.841} & 23.67 \\
BudCache & 20.000 & 147.331 & 2.29$\times$ & 80.66\% & 0.107 & 0.838 & \underline{24.02} \\
\midrule
DP Init & 20.000 & 147.137 & 2.29$\times$ & 80.51\% & 0.175 & 0.766 & 20.74 \\
\rowcolor{lightblue}
OnCache (w/o) & 20.026 & 148.326 & 2.28$\times$ & 80.91\% & 0.113 & 0.832 & 23.79 \\
\rowcolor{lightblue}
OnCache (w) & 19.901 & 148.365 & 2.28$\times$ & \textbf{81.35\%} & \textbf{0.093} & \textbf{0.854} & \textbf{24.89} \\
\bottomrule
\end{tabularx}
\end{minipage}
\end{table*}
\begin{figure*}[t]
    \centering
    \includegraphics[width=0.88\linewidth]{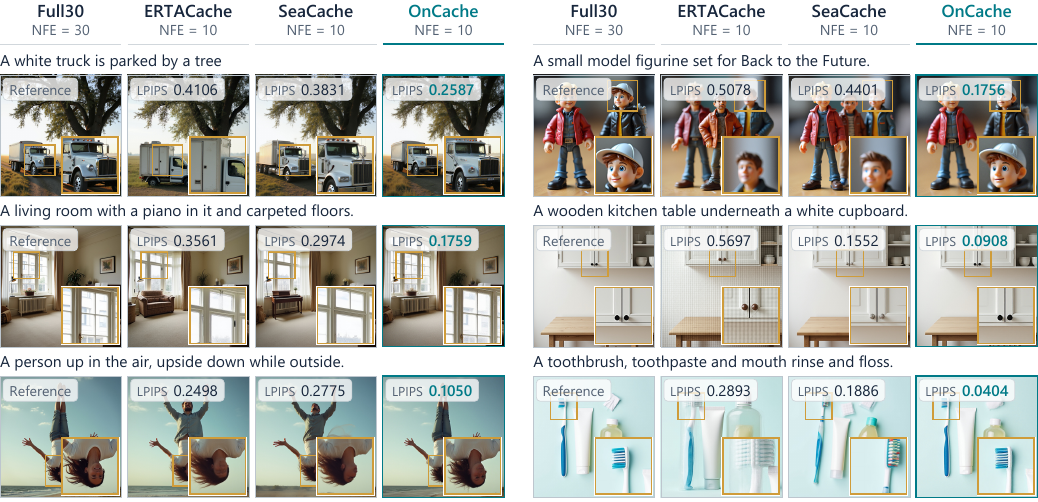}
    \par\nointerlineskip\vspace{1pt}
    {\color{black!40}\rule{.86\linewidth}{.7pt}}
    \par\nointerlineskip\vspace{1pt}
    \includegraphics[width=0.88\linewidth]{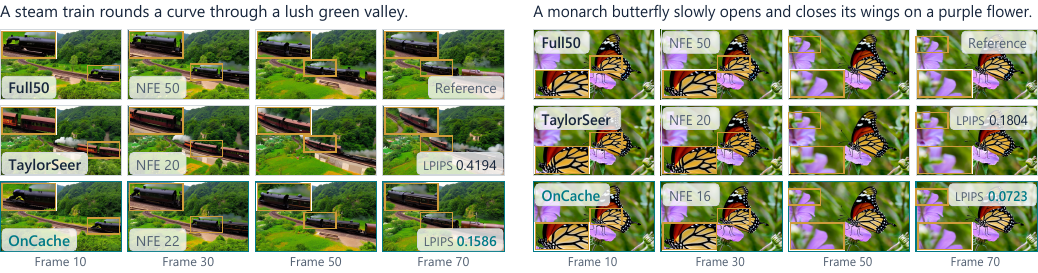}
    \vspace{-5pt}\caption{\textbf{Qualitative comparison of image and video.}
    \textbf{Upper:} six FLUX.1-dev images from COCO.
    \textbf{Lower:} four evenly spaced frames from two Wan2.1-T2V-1.3B videos. Gold boxes and insets mark shared regions. Appendix~\ref{app:qualitative} gives further image comparisons.}
    \label{fig:qualitative-native}
\end{figure*}
\subsection{Corrector for Reused Hidden States}
\label{sec:corrector}
\paragraph{Motivation.}
Cache reuse inevitably introduces approximation errors. The corrector reduces
these errors locally: on 640 same-state validation queries
(Fig.~\ref{fig:method-diagnostics}b), the frozen corrector reduces mean
relative output error by 19.5\% (95\% prompt-bootstrap interval 19.0--20.1\%). By improving predictions at cache steps, it can also mitigate error propagation through subsequent denoising steps (Fig.~\ref{fig:corrector-error-propagation}).

\paragraph{Cache-anchored hidden correction.}
With policy $\bar\pi=\pi_{\bar\theta,\bar b}$ fixed, a lightweight network predicts a hidden adjustment and applies it before the output head:
\begin{equation}
\begin{aligned}
 \Delta h_i=\mathcal C_\phi(\psi_i),\quad
 \hat v_i=\mathcal P_i\!\left(\bar h_i+
 \mathbf1_{\{a_i=0\}}\eta\Delta h_i\right).
\end{aligned}
\label{eq:correction}
\end{equation}
The inputs $\psi_i$ summarize the most recent paid residual, its first
difference from the preceding paid residual, input drift, and sampler time.
The first difference is scaled by the actual noise-level intervals. A depthwise convolution and channel gating mix the features; time-conditioned normalization and a gated feed-forward block precede a zero-initialized
projection to the hidden space (see Appendix~\ref{app:corrector} for implementation details and Sec.~\ref{sec:ablation} for component ablations).

\paragraph{Same-state supervision.}
With the policy fixed, we run the full backbone offline on cache steps. These evaluations leave the sampling
trajectory and residual history unchanged. We detach the teacher outputs
from the computation graph and use them to define targets for the hidden and final states:
\begin{equation}
 \Delta h_i^\star=\mathcal B_i(h_i,c)-\bar h_i,
 \qquad v_i^\star=\mathcal P_i(\mathcal B_i(h_i,c)).
\label{eq:corrector-target}
\end{equation}
The corrector learns the hidden representation and aligns the resulting denoiser output with full computation through the frozen output head. Using SmoothL1 loss $\operatorname{SL1}$, we write:
\begin{equation}
\begin{aligned}
 \ell_h&=\operatorname{norm}_h\!\left[\operatorname{SL1}(\Delta h_i,\Delta h_i^\star)\right],\\
 \ell_v&=\operatorname{norm}_v\!\left[\operatorname{SL1}\!\left(\mathcal P_i(\bar h_i+\Delta h_i),v_i^\star\right)\right],\\
 \mathcal L_{\mathrm{corr}}&=\mathbb E[\ell_h+\lambda_v\ell_v].
\end{aligned}
\label{eq:corrector-loss}
\end{equation}
The normalization operators use training-set hidden and output scales. We first fit the hidden loss, add output consistency with
$\lambda_v=0.1$, and then refine on a balanced mix of original and corrected trajectories. Only $\phi$ receives gradients. At inference, both modules are frozen, with the corrector compensating for cache-induced errors and the policy observing the corrected resulting trajectory.

\section{Experiments}
\label{sec:experiments}
\subsection{Experimental Settings}
\label{sec:flux-experimental-setup}
\paragraph{Models and evaluation.}
We evaluate FLUX.1-dev~\citep{fluxdevmodelcard} on 1,000 COCO captions~\citep{DBLP:conf/eccv/LinMBHPRDZ14} at $512^2$ and
$1024^2$, using 30 sampling steps and seed 42. Wan2.1-T2V-1.3B is evaluated on the full VBench~\citep{DBLP:conf/cvpr/HuangHYZS0Z0JCW24}, using five seeds per prompt (0--4), 50 sampling steps, 81 frames, and $832\times480$ resolution. 
For FLUX, we compare TeaCache~\citep{liu2025teacache},
SeaCache~\citep{chung2026seacache}, DiCache~\citep{bu2026dicache},
Spectrum~\citep{han2026spectrum}, TaylorSeer~\citep{liu2025taylorseer},
DPCache~\citep{cui2026dpcache}, BudCache~\citep{lei2026budcache}, and
ERTACache~\citep{peng2026ertacache} using public implementations.
Wan uses the same baselines except Spectrum and DPCache.
DP Init provides our static initialization. OnCache variants (with / without corrector) target mean NFE budgets of ten for FLUX and twenty for Wan.

\vspace{-8pt}
\paragraph{Metrics and timing.}
LPIPS~\citep{DBLP:conf/cvpr/ZhangIESW18}, SSIM~\citep{DBLP:journals/tip/WangBSS04},
and PSNR measure fidelity to matched full-step references.
We additionally report ImageReward~\citep{xu2023imagereward} for FLUX and VBench~\citep{DBLP:conf/cvpr/HuangHYZS0Z0JCW24} for Wan. Generation is timed on one NVIDIA H20 GPU, including online overhead, sampling, and VAE decoding. Speedup is relative to the full reference.
Appendix~\ref{app:protocol} gives software details and baseline settings.

\subsection{Main Experimental Results}
\label{sec:flux-final}
Tab.~\ref{tab:flux-cache-comparison} and~\ref{tab:wan-cache-comparison}
report the results. On both backbones, the policy alone improves all fidelity metrics over DP Init at comparable mean NFE. The full OnCache method (with corrector) achieves state-of-the-art performance on both backbones.

\vspace{-6pt}
\paragraph{Quantitative Comparison.} For image generation on FLUX, at $512^2$, OnCache (w) achieves 0.127 LPIPS, 0.857 SSIM, 25.55\,dB PSNR, and 0.989 ImageReward, outperforming all accelerated baselines on every metric. Compared with BudCache, LPIPS improves from 0.143 to 0.127 and PSNR from 24.77 to 25.55\,dB, while generation time remains close
(2.113 vs. 2.095\,s). For video generation on Wan, adding the corrector improves both reference fidelity and VBench with slightly lower mean NFE and nearly unchanged latency. At a $2.28\times$ speedup, our OnCache achieves a VBench score just 0.09\% below full-step inference.

\paragraph{Qualitative fidelity.}
Figure~\ref{fig:qualitative-native} shows that OnCache better preserves visual details in both images (upper) and videos (lower). Appendix~\ref{app:qualitative} provides more examples.

\subsection{Model Size and Computational Cost}
\label{sec:cost}
Table~\ref{tab:module-cost} shows that both modules are lightweight and add little generation time on both backbones. The policy and corrector weights total 4.42\,MB for FLUX and 2.47\,MB for Wan. Enabling both modules increases measured generation time over DP Init by 1.05\% and 0.83\%, respectively. 
Appendix~\ref{app:module-cost} details offline teacher trajectory collection and policy/corrector optimization costs for both models.

\begin{table}[h]
\centering
\footnotesize
\setlength{\tabcolsep}{2pt}
\caption{Module size and generation time on FLUX and Wan.}
\label{tab:module-cost}
\begin{tabular}{@{}llccc@{}}
\toprule
Model & Config & Params & Size (MB) & Time (s) \\
\midrule
FLUX & DP Init & -- & -- & 2.091 \\
     & + Policy & 51.7K & 0.418 & 2.098 ($+0.33\%$) \\
     & + Corrector & 1.00M & 4.003 & 2.113 ($+0.71\%$) \\
\midrule
Wan  & DP Init & -- & -- & 147.137 \\
     & + Policy & 4.36K & 0.045 & 148.326 ($+0.81\%$) \\
     & + Corrector & 0.606M & 2.424 & 148.365 ($+0.03\%$) \\
\bottomrule
\end{tabular}
\end{table}

\subsection{Generalization Analysis}
\label{sec:generalization}
\paragraph{Across resolutions.}
The policy and corrector are trained on mixed $256^2$ and $512^2$ data.
The policy uses fixed-dimensional pooled state features, while the
corrector shares channel projections and spatial convolutions across
token grids. OnCache generalizes zero-shot to the unseen $1024^2$
resolution, achieving 0.156 LPIPS without fitting or selection at that
resolution (as shown in Table~\ref{tab:flux-cache-comparison}).

\vspace{-8pt}
\paragraph{Across budgets.}
Adjusting only the operating offset allows the same frozen policy to target mean budgets of 8, 10, and 12 NFE. As computation increases, LPIPS consistently decreases while SSIM and PSNR increase (Table~\ref{tab:frozen-budget-curve}). The frozen corrector further improves all three metrics at every budget,
showing consistent benefits across the policy trajectories induced at different budgets.

\begin{table}[h]
\centering
\footnotesize
\setlength{\tabcolsep}{2pt}
\caption{Frozen modules across budgets on 200 test captions. w/o and w denote OnCache without and with the corrector.}
\label{tab:frozen-budget-curve}
\begin{tabular}{@{}cccccc@{}}
\toprule
Method & Target & NFE & LPIPS$\downarrow$ & SSIM$\uparrow$ & PSNR$\uparrow$ \\
\midrule
w/o & 8 & 8.000 & 0.2271 & 0.7541 & 21.839 \\
w & 8 & 8.000 & 0.2009 & 0.7861 & 22.512 \\
w/o & 10 & 10.000 & 0.1449 & 0.8364 & 24.770 \\
w & 10 & 10.085 & 0.1252 & 0.8602 & 25.447 \\
w/o & 12 & 12.045 & 0.1066 & 0.8808 & 26.303 \\
w & 12 & 12.030 & 0.0950 & 0.8935 & 26.747 \\
\bottomrule
\end{tabular}
\end{table}

\vspace{-8pt}
\paragraph{Across schedules (Corrector's generalization).} The corrector also improves a fixed BudCache schedule, enhancing all three fidelity metrics without retraining
(Appendix~\ref{app:frozen-compensation}).

\subsection{Ablation Studies}
\label{sec:ablation}

\begin{figure}[t]
\centering
\includegraphics[width=0.85\linewidth]{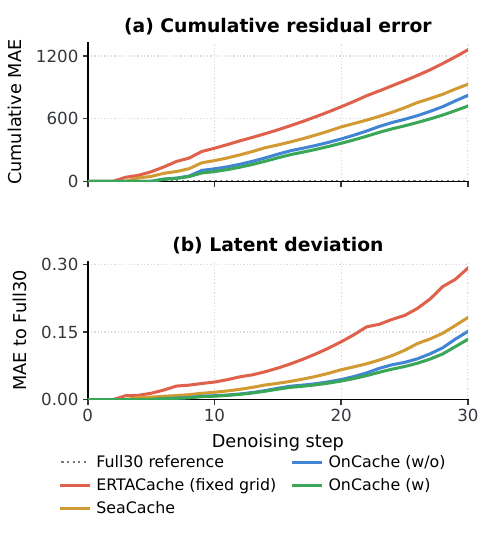}
\vspace{-5pt}
\caption{\textbf{Error propagation on FLUX.} Cumulative hidden-residual
error (top) and latent deviation (bottom) from Full30, averaged over five
prompts and five seeds at $512^2$ and 10 NFE.}
\label{fig:corrector-error-propagation}
\vspace{-5pt}
\end{figure}

\paragraph{Error propagation.}
Fig.~\ref{fig:corrector-error-propagation} compares denoising errors at
10 NFE. OnCache (w/o) already has lower cumulative hidden-residual error and
latent deviation than SeaCache and ERTACache. Adding the corrector further reduces both errors. The full method shows more gradual error growth in later denoising steps and lower terminal deviation, highlighting the effectiveness of the policy and corrector. 

\begin{figure}[t]
\centering
\includegraphics[width=0.93\linewidth]{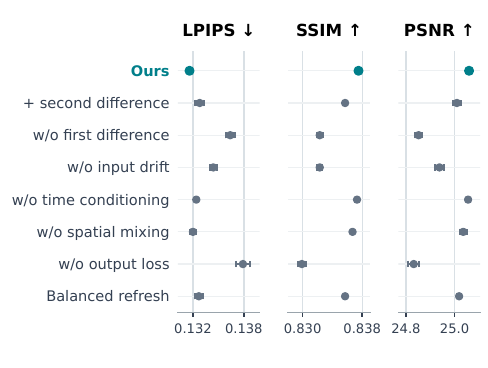}
\vspace{-5pt}
\caption{\textbf{Corrector component ablations.} Fidelity on 200 captions under the same 10/30 schedule. Points and whiskers
show the mean and sample standard deviation across three seed means.}
\label{fig:small-corrector-fidelity}
\vspace{-6pt}
\end{figure}

\paragraph{Corrector components.}
Figure~\ref{fig:small-corrector-fidelity} compares the corrector's inputs,
architecture, and supervision under a fixed training budget and refresh
schedule. Each variant is trained from scratch on 64 captions for 1,280
updates, with the backbone and policy frozen. The full design achieves the
best mean LPIPS, SSIM, and PSNR across three training seeds. Removing the
first residual difference, input drift, or output-velocity supervision causes
the largest fidelity drops. Learned time conditioning and spatial mixing
provide smaller gains, while adding a second difference or balanced trajectory
refresh gives no improvement under this short training schedule.
Appendix~\ref{app:small-corrector-components} details the training protocol
and controls.

\vspace{2pt}
\section{Conclusion}
We presented OnCache, a dynamic caching framework that allocates computation online across samples and timesteps under average budgets. Paired-continuation supervision guides refresh decisions through their downstream effects on final-output fidelity and total computation, while a lightweight neural corrector mitigates feature-reuse errors and their propagation. Both modules are trained offline with a frozen backbone and add minimal inference overhead. Experiments on image and video models demonstrate state-of-the-art efficiency--quality trade-offs.

\clearpage
\section*{LLM Usage Statement}
ChatGPT assisted with literature organization, writing, and LaTeX formatting. The authors take full responsibility for the scientific content and supporting evidence.

{\small
\bibliographystyle{ieeenat_fullname}
\bibliography{ref}
}
\input{appendix}
\end{document}

%% file: appendix.tex
\clearpage
\appendix
\onecolumn
\setcounter{page}{1}
\begingroup
\renewcommand{\twocolumn}[1][]{#1}
\maketitlesupplementary
\par
\endgroup
\begingroup
\noindent\begin{minipage}{\textwidth}
{\Large\bfseries Appendix Guide}\par
\vspace{0.4em}
{\small Evaluation details, implementation, and additional analysis of OnCache.}
\par\vspace{0.7em}
{\color{cvprblue}\rule{\linewidth}{1pt}}
\par\vspace{0.5em}
\newcommand{\appendixguidesubentry}[2]{%
  \noindent\hspace*{1.15em}%
  \begin{minipage}[t]{\dimexpr\linewidth-4.5em\relax}
    \small\hyperref[#1]{\makebox[2.6em][l]{{\color{cvprblue}\bfseries\ref*{#1}}}{\color{black!90}\bfseries\renewcommand{\texorpdfstring}[2]{##2}\nameref*{#1}}}%
    {\color{black!75}\enspace---\enspace#2}%
  \end{minipage}%
  \hfill\hyperref[#1]{{\small\color{cvprblue}p.~\pageref*{#1}}}\par
  \vspace{0.2em}
}
\newcommand{\appendixguideentry}[3]{%
  \noindent
  \hyperref[#1]{{\color{cvprblue}\bfseries\ref*{#1}}\quad\textbf{\nameref*{#1}}}%
  \hfill\hyperref[#1]{{\small\color{cvprblue}p.~\pageref*{#1}}}\par
  \vspace{0.3em}{\small\color{black!75}#2\par}
  \vspace{0.35em}
  #3
  \vspace{0.3em}{\color{cvprblue!20}\hrule height 0.4pt}
  \vspace{0.65em}
}
\appendixguideentry{app:protocol}{%
  Hardware, software, metrics, and baseline settings for FLUX and Wan.}{}
\appendixguideentry{app:policy}{%
  DP profiling, paired-value fitting and refinement, and online inference.}{%
  \appendixguidesubentry{app:dp-init}{Residual profiling and anchor schedules.}
  \appendixguidesubentry{app:value-fit}{Policy fitting, refinement, and deployment.}
}
\appendixguideentry{app:corrector}{%
  Hidden-state correction with a frozen policy, from architecture to inference.}{%
  \appendixguidesubentry{app:corrector-architecture}{History features and network design.}
  \appendixguidesubentry{app:corrector-training}{Teacher targets, staged training, and closed-loop inference.}
}
\appendixguideentry{app:policy-ablations}{
  Schedule assignment, target/input ablations, and cross-resolution generalization.}{
  \appendixguidesubentry{app:frozen-policy-routing}{Permutation controls at fixed computation.}
  \appendixguidesubentry{app:small-policy-components}{Continuation costs, targets, and input ablations.}
  \appendixguidesubentry{app:training-resolution}{Single- and mixed-resolution fits over four rounds.}
}
\appendixguideentry{app:corrector-ablations}{%
  Compensation rules, components, training objectives, and transfer.}{%
  \appendixguidesubentry{app:frozen-compensation}{Compensation at a shared refresh schedule.}
  \appendixguidesubentry{app:small-corrector-components}{Training protocol and component controls.}
  \appendixguidesubentry{app:corrector-loss-objectives}{Hidden, velocity, and joint loss controls.}
  \appendixguidesubentry{app:corrector-transfer}{Frozen correction on BudCache and DiCache.}
}
\appendixguideentry{app:module-cost}{
  Offline collection/optimization costs and 50-step comparison with training-based method ReCache.}{
  \appendixguidesubentry{app:policy-cost}{Recorded data collection and fitting time.}
  \appendixguidesubentry{app:training-cache-comparison}{50-step comparison with ReCache and offline costs.}
}
\appendixguideentry{app:related}{
  Diffusion acceleration, feature reuse, and sequential policy learning.}{
  \appendixguidesubentry{app:related-reuse}{Block/token reuse methods and trajectory-adaptive caching.}
  \appendixguidesubentry{app:related-prediction}{Feature forecasting, correction, and budgets.}
  \appendixguidesubentry{app:related-sequential}{Learning from sequential decisions and outcomes.}
}
\appendixguideentry{app:qualitative}{
  Additional FLUX images and Wan videos compared with other caching baselines.}{}
\end{minipage}
\endgroup
\space
\twocolumn
\section{Evaluation Protocol and Reproducibility}
\label{app:protocol}

\begin{table}[h]
\centering
\small
\setlength{\tabcolsep}{6pt}
\caption{Hardware / software for the main generation experiments.}
\label{tab:experiment-environment}
\begin{tabular}{@{}lcc@{}}
\toprule
Component & FLUX & Wan \\
\midrule
GPU & \multicolumn{2}{c}{NVIDIA H20} \\
Python & \multicolumn{2}{c}{3.11.18} \\
CUDA & \multicolumn{2}{c}{13.0} \\
PyTorch & \multicolumn{2}{c}{2.14.1+cu130} \\
Diffusers & 0.41.0 & 0.33.1 \\
Transformers & 5.20.2 & 4.49.0 \\
NumPy & 2.2.6 & 1.26.4 \\
\bottomrule
\end{tabular}
\end{table}

\begin{table*}[t]
\centering
\begin{minipage}{0.8\textwidth}
\centering
\footnotesize
\setlength{\tabcolsep}{3pt}
\renewcommand{\arraystretch}{1.05}
\caption{Baseline settings for FLUX.1-dev and Wan2.1-T2V-1.3B. Shown NFE values are means from Tables~\ref{tab:flux-cache-comparison} and \ref{tab:wan-cache-comparison}. OnCache (w/o) and (w) share the policy without and with the corrector; -- denotes a method not compared.}
\label{tab:frozen-baseline-settings}
\vspace{-5pt}
\begin{tabularx}{\textwidth}{@{}l>{\centering\arraybackslash}X>{\centering\arraybackslash}X@{}}
\toprule
Method & FLUX: $512^2$, 30 steps & Wan: 50 steps \\
\midrule
DP Init & Static schedule, 10 NFE & Static schedule, 20 NFE \\
OnCache (w/o) & $b=-0.559$; no corrector & $b=3.93$; no corrector \\
OnCache (w) & Same policy, with corrector & Same policy, with corrector \\
TeaCache & Adaptive gate; 10.033 NFE & Adaptive gate; 23.741 NFE \\
SeaCache & Spectral gate; 10.003 NFE & Spectral gate; 20.007 NFE \\
DiCache & One-block probe; 10.006 NFE & One-block probe; 23.718 NFE \\
TaylorSeer & Block-wise forecast; 10 NFE & Block-wise forecast; 20 NFE \\
Spectrum & Chebyshev forecast; 10 NFE & -- \\
DPCache & DP-optimized schedule; 10 NFE & -- \\
ERTACache & Alignment and rectification; 10 NFE & Alignment and rectification; 24 NFE \\
BudCache & Stage 1 schedule; 10 NFE & Stage 1 schedule; 20 NFE \\
\bottomrule
\end{tabularx}
\end{minipage}
\end{table*}

Table~\ref{tab:experiment-environment} summarizes the hardware / software used for the main FLUX and Wan experiments. The code and experiment configurations will be released. For our fidelity measurements, each accelerated sample and its full-step reference use the same prompt and initial noise. LPIPS uses AlexNet on uncropped native-resolution images; SSIM and PSNR use scikit-image.
Table~\ref{tab:frozen-baseline-settings} summarizes the main method
settings for both FLUX and Wan main experiments. 

\section{Policy Implementation and Training}
\label{app:policy}
This section details the policy implementation, training, and online inference.

\subsection{DP Profiling and Search}
\label{app:dp-init}
\paragraph{Profiling.}
For both FLUX and Wan, we run full-evaluation trajectories on profiling
prompts and subsample the block-stack residuals at each step to obtain compact summaries $\Xi_{p,t}$. Equation~\ref{eq:dp-mismatch} in the main paper compares summaries from a candidate refresh step and a later step, estimating the
mismatch from residual reuse across prompts. These estimates provide the proxy costs for the DP search.

\paragraph{Search.}
The interval cost $C(i,j)$ sums the mismatch proxy over steps between
successive refreshes $i$ and $j$. DP minimizes the sum of these interval
costs subject to the budget and boundary constraints. The profiler supplies
boundary paths and admissible intervals $\mathcal E_{\mathrm{DP}}$, so the
predecessors of $j$ are
$\mathcal I_j=\{i:0\le i<j,\ (i,j)\in\mathcal E_{\mathrm{DP}}\}$.
We fill the table in increasing order of $k$ and record each improving
predecessor. Backtracking recovers the refresh steps; an infinite terminal
cost indicates an infeasible budget. Given the interval costs, the search
takes $O(K|\mathcal E_{\mathrm{DP}}|)$ time and $O(KN)$ storage.

\paragraph{Detailed schedule.}
For 30-step FLUX generation, the retained 10-evaluation DP anchor uses indices:
\begin{equation*}
 \mathcal F_{\mathrm{FLUX}}=\{0,1,2,3,4,5,8,15,23,29\}.
\end{equation*}
The same anchor is used at both $512^2$ and $1024^2$.

For 50-step Wan generation, the 20-NFE DP schedule is:
\begin{equation*}
\begin{aligned}
 \mathcal F_{\mathrm{Wan}}=\{&0,1,2,3,4,5,6,7,9,11,14,\\
 &17,20,24,29,33,37,41,45,48\}.
\end{aligned}
\end{equation*}
These schedules initialize data collection, serve as DP Init controls, and
provide the policy's time prior. The deployed policy chooses refresh steps online.
Algorithm~\ref{alg:dp-search} summarizes the dynamic programming search and traceback.

\begin{algorithm}[h]
\caption{DP search for an initial refresh schedule}
\label{alg:dp-search}
\small
\begin{algorithmic}[1]
\Require Steps $N$, budget $K$, interval costs $C$, predecessors $\mathcal I_j$
\Require Initial table $D_0$ with boundary paths; terminal boundary conditions and costs
\Ensure A $K$-evaluation schedule, or infeasibility
\State $D\gets D_0$; initialize records for the boundary paths
\For{$k=1,\ldots,K$}
    \For{$j=0,\ldots,N-1$}
        \For{$i\in\mathcal I_j$}
            \If{$D[k-1,i]+C(i,j)<D[k,j]$}
                \State $D[k,j]\gets D[k-1,i]+C(i,j)$
                \State Record $i$ as the predecessor of $(k,j)$
            \EndIf
        \EndFor
    \EndFor
\EndFor
\State Select the lowest-cost admissible terminal path with $K$ evaluations, including boundary costs
\If{no finite admissible terminal path exists}
    \State \Return infeasible
\EndIf
\State Backtrack through the recorded predecessors to an initial boundary path
\State \Return the boundary path and recovered refresh steps
\end{algorithmic}
\end{algorithm}

\subsection{Policy Training and Inference}
\label{app:value-fit}
Policy learning uses paired continuations to estimate the terminal fidelity
gain and total computation change of a refresh. The backbone remains frozen. We first fit an initial score using states from the DP schedule and fit updated paired values on states visited by the
resulting policy.

\subsubsection{Initial Paired-Value Model}
At each selected state, the two continuations start from the same latent,
cache, and history, take opposite actions, and then follow the same frozen
collection policy on their respective trajectories. Eq.~\ref{eq:targets}
defines their compute-minus-cache fidelity gain $\mathbf g_q$ and total NFE
change $d_q$. The predictor $f_W$ in Eq.~\ref{eq:value-regression} fits these
targets from state summaries, time features, and their interactions. State
features and targets are normalized using fitting data; robust weighting
down-weights large-residual outliers, while query weights balance sample
contributions across groups such as prompts or training conditions.
The initial score combines these predictions with the DP time prior:
\begin{equation}
 s_\theta(z_i)=s_{\mathrm{DP}}(i)
 +\frac{\mathbf w^\top\hat{\mathbf g}_i}{\max(\hat d_i,\epsilon_d)}.
\label{eq:value-index}
\end{equation}
The DP-dependent term includes a fixed initial offset and adds a positive
margin at DP refresh steps and a negative margin at other eligible steps. The gain
weights $\mathbf w$ include the scale of the prediction term, and the positive
floor $\epsilon_d$ stabilizes the predicted cost in the ratio (the measured
NFE change $d_i$ can still be zero or negative). Using this benefit-per-unit-cost
ratio provides a scale-consistent initial ranking around the static DP anchor
before switching to linear net-value refinement in later rounds. The shared
offset $b$ in Eq.~\ref{eq:score} separately controls the overall refresh tendency.

\subsubsection{Refreshed Paired Values}
\label{app:adaptive-value-fit}
Once the policy moves beyond the static DP anchor, Eq.~\ref{eq:policy-update}
refines the score using the linear net-value predictor in
Eq.~\ref{eq:policy-value} fitted on states visited by the current policy. On
the right-hand side, $s_\theta$ and $b$ belong to the fixed collection policy;
the arrow replaces its score function with the updated one for the next round.
Because each round refits the predictor on an updated visited-state
distribution, the gain and NFE weights absorb the new prediction scale, and
the symmetric clip bound keeps score updates properly bounded across rounds.
At inference (Algorithm~\ref{alg:inference}), the final policy and offset $b$
remain fixed, with an optional consecutive-reuse limit $L$ safeguarding against
overly long reuse intervals.

\begin{algorithm}[t]
\caption{OnCache decision at a cache-eligible step}
\label{alg:inference}
\small
\begin{algorithmic}[1]
\Require Frozen policy score $s_\theta$, offset $b$, reuse limit $L$, latent $x_i$, conditioning $c$, valid residual $R_j$, history $\mathcal H_i$; corrector $\mathcal C_\phi$ and strength $\eta$ when enabled
\State $h_i\gets\mathcal E_i(x_i,c)$
\State $z_i\gets\varphi(h_i,c,i,\mathcal H_i)$
\State $a_i\gets\mathbf1_{\{s_\theta(z_i)+b\geq0\}}$
\If{$L$ consecutive cache actions precede $i$}
    \State $a_i\gets1$
\EndIf
\If{$a_i=1$}
    \State $\bar h_i\gets\mathcal B_i(h_i,c)$
    \State $R_i\gets\bar h_i-h_i$; update residual history
\Else
    \State $\bar h_i\gets h_i+R_j$
    \If{corrector enabled}
        \State Construct $\psi_i$ from $h_i$, sampler time, and residual history
        \State $\bar h_i\gets\bar h_i+\eta\mathcal C_\phi(\psi_i)$
    \EndIf
\EndIf
\State $v_i\gets\mathcal P_i(\bar h_i)$
\State $x_{i+1}\gets\mathcal S_i(x_i,v_i)$
\State Record $a_i$ and realized NFE
\end{algorithmic}
\end{algorithm}

\section{Corrector Implementation and Training}
\label{app:corrector}
The corrector estimates a hidden-state correction for a cached feature. It is trained offline with the policy frozen, then applied to the reused representation before the output head.

\subsection{Causal Inputs and Architecture}
\label{app:corrector-architecture}
At a cache step $i$, let $j$ denote the most recent full evaluation. The
input $\psi_i$ combines the cached residual $R_j$, its first difference
from the preceding full evaluation, and input drift $h_i-h_j$. The first
difference is divided by the noise-level gap between those evaluations and
scaled by the gap from the latest evaluation to the current step; it is
masked only during early caching steps before two full evaluations have
accumulated. Cache age, sampling progress, and feature statistics provide
additional conditioning.

A shared projection compresses these features, which are fused by a depthwise spatial convolution with channel gating. Time-conditioned
normalization, a gated feed-forward block, and a zero-initialized output
projection produce the hidden correction $\Delta h_i$. This zero
initialization provides an identity-preservingstarting point that matches the unperturbed cached representation and prevents early random perturbations from destabilizing the denoising trajectory.

\subsection{Supervision and Deployment}
\label{app:corrector-training}
\label{app:corrector-inference}
Offline teacher queries on policy-guided cache states provide the hidden
and output targets in Eq.~\ref{eq:corrector-target}, without changing the
trajectory or cache. In Eq.~\ref{eq:corrector-loss}, $\operatorname{norm}_h$ divides the token/channel-averaged hidden loss by cache age and the mean
hidden-target magnitude. $\operatorname{norm}_v$ scales the output difference after SmoothL1 using the mean full-minus-cached output magnitude. Both normalization scales are computed solely from the training split.

Training follows a staged schedule: it first fits the hidden-state loss, then adds output-velocity consistency ($\lambda_v=0.1$), and finally refines on an equal (1:1) mix of original and corrected trajectories to align training with closed-loop rollout dynamics. The corrector is trained offline on the collected teacher-trajectory queries, with correction strength fixed at $\eta=1$ across our evaluations. At inference, the frozen corrector adjusts cache steps, only full evaluations update the residual history, and the frozen policy observes the corrected trajectory. As shown in the component ablations in Sec.~\ref{sec:ablation} (Fig.~\ref{fig:small-corrector-fidelity}), the full design achieves the best mean LPIPS, SSIM, and PSNR across three training seeds, with the first residual difference, input drift, and output-velocity supervision providing the primary fidelity gains. Detailed training protocols and controls are reported in Appendix~\ref{app:small-corrector-components}, and loss objectives at matched updates are compared in Appendix~\ref{app:corrector-loss-objectives}.

\section{Policy Experiments and Ablations}
\label{app:policy-ablations}
We first verify that the policy's cache allocation adapts both per-sample computation and refresh timing to individual trajectories (Appendix~\ref{app:frozen-policy-routing}). We then isolate the contributions of learned continuation-cost targets and state-history inputs using compact single-fit learners (Appendix~\ref{app:small-policy-components}). Finally, we ablate single-/mixed-resolution training to examine cross-resolution generalization (Appendix~\ref{app:training-resolution}).

\subsection{Schedule Assignment with the Final Policy}
\label{app:frozen-policy-routing}
We verify that our final frozen policy ($b=-0.559$, without corrector) adapts both per-sample computation and refresh placement to individual trajectories. The evaluation uses 200 test captions at $512^2$ resolution with paired Full30 references (seed 42, 30 sampling steps, guidance 3.5), disjoint from all teacher-trajectory and fitting sets.

We compare the policy's own schedules against two seeded permutations that
preserve the schedule multiset and all 1,973 full evaluations (mean NFE
9.865): an \emph{all-prompt permutation} that reassigns schedules across all
captions (disrupting both per-sample budget and refresh timing), and a
\emph{within-NFE permutation} that permutes schedules only among captions
sharing the same NFE (isolating refresh timing while keeping each prompt's
computation fixed).

\begin{table}[t]
\centering
\small
\setlength{\tabcolsep}{3pt}
\caption{Final-policy schedule assignment on 200 $512^2$ captions at
matched total computation. Bold marks the best scores.}
\label{tab:frozen-policy-routing}
\resizebox{0.75\linewidth}{!}{%
\begin{tabular}{@{}lcccc@{}}
\toprule
Assignment & NFE & LPIPS$\downarrow$ & SSIM$\uparrow$ & PSNR$\uparrow$ \\
\midrule
Our Policy & 9.865 & \textbf{0.1482} & \textbf{0.8319} & \textbf{24.70} \\
All-prompt permutation & 9.865 & 0.1676 & 0.8124 & 23.72 \\
Within-NFE permutation & 9.865 & 0.1593 & 0.8203 & 23.84 \\
\bottomrule
\end{tabular}%
}
\vspace{-10pt}
\end{table}

As shown in Table~\ref{tab:frozen-policy-routing}, both permutations degrade
every fidelity metric at identical total computation, with the all-prompt permutation suffering the largest drop ($+0.0194$ LPIPS, $-0.0195$ SSIM, and
$-0.98$\,dB PSNR). Even under the stricter within-NFE control where every
caption retains its exact NFE, our policy's placement reduces LPIPS by
$0.0111$ (95\% CI $[-0.0131,-0.0097]$), improves SSIM by $0.0116$ ($[0.0102,0.0132]$), and raises PSNR by $0.86$ ($[0.7629,0.9714]$),
confirming that the policy effectively adapts both computation allocation across different prompts and
refresh timing to each trajectory.

\subsection{Paired-Value Targets and Policy Inputs}
\label{app:small-policy-components}

\paragraph{Measured continuation cost.}
In sequential caching, refreshing at step $i$ changes future states and can either trigger or save subsequent full evaluations, so its true computation cost is not a constant $+1$ NFE. To verify this motivation, we collect seven paired action queries per caption from 64 teacher-trajectory captions and an additional 16-caption diagnostic set. Each pair shares the same prompt, initial noise, action prefix, and pre-action cache state, takes opposite actions at the queried step, and then follows the same frozen final policy without the corrector. Query steps are chosen from pre-action decision margins before endpoint fidelity is observed, and all 560 pairs pass prefix, image, and execution-count checks. 
Table~\ref{tab:paired-policy-cost} reports the resulting trajectory-level NFE
difference $d_i$ from Eq.~\ref{eq:targets}. Across the 448 fitting pairs, 206
(46.0\%) have $d_i\ne1$, including 201 zero-cost cases where an immediate
refresh saves a later full evaluation, yielding a mean cost of $0.556$
(ranging from $-1$ to $2$). The 16-caption diagnostic set shows the same
pattern (mean $0.714$, with 44 of 112 pairs having $d_i\ne1$). These
measurements motivate learning the continuation cost $d_i$ rather than
assuming a fixed unit cost.

\begin{table}[h]
\centering
\small
\setlength{\tabcolsep}{4pt}
\caption{Measured compute-minus-cache continuation NFE differences ($d_i$),
accounting for the current and all subsequent actions.}
\vspace{-5pt}
\label{tab:paired-policy-cost}
\resizebox{\columnwidth}{!}{%
\begin{tabular}{@{}lcccccc@{}}
\toprule
Dataset & Pairs & $d=-1$ & $d=0$ & $d=1$ & $d=2$ & Mean $d$ \\
\midrule
Teacher set (64) & 448 & 1 & 201 & 242 & 4 & 0.556 \\
Diagnostic set (16) & 112 & 0 & 38 & 68 & 6 & 0.714 \\
\bottomrule
\end{tabular}%
}
\vspace{-15pt}
\end{table}

\paragraph{Compact target and input controls.}
To isolate the contributions of the learned cost target and state-history
inputs under a controlled single-fit budget, we train four compact
paired-value learners on the same 64-caption teacher-trajectory set via
regularized least squares with fitting-set normalization. The \emph{Paired
learner (Ours)} uses full state-time features to predict both normalized
fidelity gains and the measured continuation cost $d_i$ in
Eq.~\ref{eq:targets}. The \emph{Unit cost target} control replaces only the
cost target with a constant $1$. To test input necessity, \emph{Time inputs
only} retains only sampler and time features, while \emph{History/drift removed} discards residual, history, and drift features (and their
interactions) while keeping current hidden summaries, action counters, and time embedding features.

All four learners share the same ten-evaluation BudCache time prior
$s_{\mathrm{prior}}$ and a bounded softsign score mapping:
\begin{equation}
 u=\hat q-\lambda\hat d-s_{\mathrm{prior}},\qquad
 s=s_{\mathrm{prior}}+\alpha\frac{u}{1+|u|}+b,
\label{eq:small-policy-score}
\end{equation}
where $\hat q$ and $\hat d$ denote the mean predicted normalized gain and
cost, and $(\lambda, \alpha)$ are fixed scaling coefficients shared across all
variants. These compact single-fit models serve strictly as controlled
lightweight ablations. All learners are frozen and evaluated on the same
200 $512^2$ test captions.

\begin{table}[h]
\centering
\small
\setlength{\tabcolsep}{3pt}
\caption{Controlled target and input ablations fitted on the 64-caption
teacher-trajectory set and evaluated on 200 $512^2$ test captions without the
corrector. Bold marks the best scores.}
\label{tab:small-policy-components}
\vspace{-5pt}
\resizebox{0.8\columnwidth}{!}{%
\begin{tabular}{@{}lcccc@{}}
\toprule
Learner & NFE & LPIPS$\downarrow$ & SSIM$\uparrow$ & PSNR$\uparrow$ \\
\midrule
Paired learner (Ours) & 10.020 & \textbf{0.1425} & \textbf{0.8285} & \textbf{24.635} \\
Unit cost target & 9.995 & 0.1491 & 0.8219 & 24.281 \\
Time inputs only & 10.000 & 0.1452 & 0.8255 & 24.573 \\
History/drift removed & 10.000 & 0.1452 & 0.8255 & 24.573 \\
\bottomrule
\end{tabular}%
}\vspace{-5pt}
\end{table}

Table~\ref{tab:small-policy-components} shows that our paired learner achieves
the best fidelity across all metrics at essentially the same computation
($10.020$ vs.\ $9.995$--$10.000$ NFE). Replacing the measured continuation
cost with a constant unit target degrades every metric at virtually
identical NFE: LPIPS worsens by $0.00662$ (95\% CI $[0.00512,0.00847]$),
SSIM drops by $0.00666$ ($[-0.00847,-0.00514]$), and PSNR drops by
$0.3541$ ($[-0.4204,-0.2967]$) across 2,000 paired prompt-bootstrap resamples, confirming that modeling downstream NFE consequences is essential. 
Furthermore, both input-reduced controls lack the trajectory-specific signals needed to depart from the time prior, collapsing identically to the static BudCache schedule and producing identical images on all 200 captions. By leveraging residual history and input drift, our paired learner outperforms both collapsed controls, demonstrating the necessity of both state-history features and
paired-continuation cost supervision.

\subsection{Training Resolution Generalization}
\label{app:training-resolution}
To evaluate how training resolution affects cross-grid generalization (complementing Sec.~\ref{sec:generalization}), we compare single- and mixed-resolution training using a compact, corrector-free policy trained over four iterative rounds. Using the same 64 teacher-trajectory captions throughout, we test three training regimes: $256^2$-only, $512^2$-only, and a balanced 1:1 split of 32 captions at each resolution (\emph{Mix}). Starting from a common ten-evaluation BudCache prior in round~1, each subsequent round collects 448 fresh paired targets along its current calibrated policy and refits the head via regularized least squares using the softsign update in Eq.~\ref{eq:small-policy-score} (1,792 paired targets in total per regime). 
After each round, a shared offset is calibrated toward a mean budget of 10~NFE using 16 calibration captions. All weights, normalizers, and offsets from the final fourth-round policy are then frozen and evaluated across $256^2$, $512^2$, and the unseen $1024^2$ grid on 64 held-out test captions (30 steps, seed 42, compared to Full30 references).

\begin{table}[h]
\centering
\small
\setlength{\tabcolsep}{4pt}
\caption{Cross-resolution generalization of the policy after four training rounds on 64 captions. Bold marks the best fidelity.}
\vspace{-5pt}
\label{tab:training-resolution-depth}
\resizebox{0.7\columnwidth}{!}{%
\begin{tabular}{@{}llcccc@{}}
\toprule
Test & Fit & NFE & LPIPS$\downarrow$ & SSIM$\uparrow$ & PSNR$\uparrow$ \\
\midrule
\multirow{3}{*}{$256^2$} & $256^2$ & 9.594 & \textbf{0.1453} & \textbf{0.8032} & 22.652 \\
 & $512^2$ & 9.578 & 0.1486 & 0.8005 & 22.406 \\
 & Mix & 9.594 & 0.1466 & 0.8019 & \textbf{22.727} \\
\midrule
\multirow{3}{*}{$512^2$} & $256^2$ & 10.172 & 0.1592 & 0.8297 & 24.712 \\
 & $512^2$ & 10.203 & \textbf{0.1517} & \textbf{0.8354} & 24.804 \\
 & Mix & 10.156 & 0.1520 & 0.8348 & \textbf{24.849} \\
\midrule
\multirow{3}{*}{$1024^2$} & $256^2$ & 11.016 & 0.1786 & 0.8446 & 25.161 \\
 & $512^2$ & 10.828 & 0.1663 & 0.8542 & 25.931 \\
 & Mix & 10.938 & \textbf{0.1503} & \textbf{0.8671} & \textbf{26.732} \\
\bottomrule
\end{tabular}%
}
\end{table}

Table~\ref{tab:training-resolution-depth} shows that mixed-resolution training delivers the strongest cross-resolution generalization without sacrificing in-domain fidelity. At both $256^2$ and $512^2$, \emph{Mix} achieves the highest PSNR while matching the in-domain single-resolution fits in LPIPS and SSIM (within 0.0013 and 0.0003 LPIPS at $256^2$ and $512^2$, respectively) at comparable NFE. 
On the unseen $1024^2$ grid, single-resolution policies degrade markedly, whereas \emph{Mix} achieves the best fidelity across all three metrics (0.1503 LPIPS, 0.8671 SSIM, and 26.732\,dB PSNR) at virtually the same computation (10.938 NFE vs.\ 11.016 and 10.828). Across 2,000 paired prompt-bootstrap resamples, \emph{Mix} reduces LPIPS significantly over both $256^2$-only (95\% CI $[-0.04057, -0.01693]$) and $512^2$-only ($[-0.02391, -0.00934]$). Exposing the policy to mixed $256^2$ and $512^2$ trajectories therefore prevents overfitting to a single spatial scale and directly supports the zero-shot $1024^2$ generalization in Sec.~\ref{sec:generalization}.

\section{Corrector Experiments and Ablations}
\label{app:corrector-ablations}
We compare compensation methods at matched computation, detail the corrector's component and loss ablations, and evaluate transfer to other caching methods.

\subsection{Correction at Matched Computation}
\label{app:frozen-compensation}

\paragraph{Correction comparison.}
While several caching methods incorporate feature compensation, existing designs rely on hand-crafted analytical rules rather than learned neural correction. For example, ERTACache~\citep{peng2026ertacache} applies
fixed-grid residual rectification, and forecasting methods such as
TaylorSeer~\citep{liu2025taylorseer}
extrapolate block-wise features via fixed Taylor expansions. Such analytical heuristics cannot adapt to trajectory-specific input drift or align reused hidden states with the denoiser's output velocity. In contrast, our corrector is a lightweight neural module trained with
same-state teacher supervision to predict hidden-state adjustments from causal residual differences, input drift, and sampler time. To isolate compensation quality from schedule selection, we fix every accelerated method's full-evaluation steps to $(0,1,2,3,4,6,9,12,19,25)$ (the 10-NFE
BudCache~\citep{lei2026budcache} schedule) and evaluate all methods on the
same 200 $512^2$ test captions in
Table~\ref{tab:frozen-compensation}. ERTACache uses fixed-grid rectification calibrated on 64 separate captions, while TaylorSeer retains its native block-wise forecasting functions.

\begin{table}[h]
\centering
\small
\setlength{\tabcolsep}{3pt}
\caption{Frozen compensation under a shared 10-NFE schedule on 200 prompts at $512^2$. Latency on one A100 GPU includes VAE decoding. Bold marks the best fidelity scores.}
\label{tab:frozen-compensation}
\resizebox{\columnwidth}{!}{%
\begin{tabular}{@{}lccccc@{}}
\toprule
Method & NFE & Latency (s)$\downarrow$ & LPIPS$\downarrow$ & SSIM$\uparrow$ & PSNR$\uparrow$ \\
\midrule
Full30 & 30 & 4.8956 & -- & -- & -- \\
\midrule
BudCache & 10 & 1.8980 & 0.1452 & 0.8255 & 24.57 \\
+ ERTACache & 10 & 1.9878 & 0.1371 & 0.8303 & 24.63 \\
+ TaylorSeer (order 1) & 10 & 2.0905 & 0.1303 & 0.8281 & 23.88 \\
+ TaylorSeer (order 2) & 10 & 2.1597 & 0.1805 & 0.7871 & 22.47 \\
+ Our corrector & 10 & 1.9392 & \textbf{0.1286} & \textbf{0.8480} & \textbf{25.20} \\
\bottomrule
\end{tabular}%
}\vspace{-10pt}
\end{table}

\paragraph{Quality and online cost.}
As shown in Table~\ref{tab:frozen-compensation}, our corrector achieves the
best fidelity across all three metrics while incurring the lowest latency
among all compensation methods. Relative to uncompensated residual reuse
(BudCache), it improves LPIPS by $0.0166$, SSIM by $0.0225$, and PSNR by
$0.63$\,dB with only $41.16$\,ms additional online latency ($+2.17\%$),
and consistently outperforms ERTACache in both fidelity and speed ($48.6$\,ms
faster).

Compared with native block-wise TaylorSeer (order~1), our corrector saves
$151.33$\,ms per image ($7.24\%$) while delivering substantial gains in
SSIM ($+0.01989$, 95\% CI $[0.01566,0.02395]$) and PSNR ($+1.32014$\,dB,
$[1.14018,1.50157]$\,dB), alongside a lower mean LPIPS ($-0.00166$,
$[-0.00568,0.00177]$) across 2,000 paired prompt-bootstrap resamples.
Compared with order-2 TaylorSeer, which degrades under this irregular
refresh schedule, our corrector saves $220.58$\,ms ($10.21\%$) and
outperforms it by a wide margin across every metric.

\subsection{Corrector Ablation Setup}
\label{app:small-corrector-components}
Figure~\ref{fig:small-corrector-fidelity} in Sec.~\ref{sec:ablation}
reports the component ablations. Each variant is trained from scratch on teacher trajectories from 64
captions with eight same-state teacher queries per caption. FLUX and the policy are frozen. Training lasts ten
epochs (1,280 optimizer updates), with four full-grid queries per batch,
learning rate $10^{-4}$, and output-loss weight 0.1 except for the
hidden-only control. Paired variants share their initialization for each of
three fixed training seeds. We evaluate the final epoch on 200 captions under
the same fixed ten-evaluation BudCache schedule.

\vspace{-5pt}
\paragraph{Controls.}
The second-difference control adds the second temporal difference and its availability flag; the first-difference control removes the first difference and its flag. The drift control removes current input drift, while the time control removes learned time conditioning but retains physical noise-level scaling. The spatial control removes depthwise convolution and channel gating together. The hidden-only control removes the output loss but retains hidden-state supervision. Balanced refresh replaces four of each caption's eight original queries with corrected-rollout queries matched by timestep and cache age. Provider fitting and the 192 refreshed fitting rollouts are shared preparation costs, separate from the final student's 1,280-update budget. The deployed corrector uses the longer staged procedure in Appendix~\ref{app:corrector}.

\FloatBarrier
\subsection{Loss Objectives at Matched Updates}
\label{app:corrector-loss-objectives}
Hidden regression anchors the correction to the teacher representation, while velocity supervision aligns the output consumed by the sampler. The frozen output head is many-to-one, so these objectives impose different constraints on the hidden correction. 
We fix the deployed architecture and the same 64 teacher-trajectory dataset of Appendix~\ref{app:small-corrector-components}, and compare $w_h\ell_h+w_v\ell_v$ with weights $(1,0)$, $(0,0.1)$, and $(1,0.1)$.
Each student starts from the same random initialization and query order for each of three fixed seeds, using AdamW with learning rate $10^{-4}$, four queries per batch, and ten epochs (1,280 updates). All variants are
evaluated directly at the final epoch with $\eta=1$ on the same 200
$512^2$ captions under the fixed ten-evaluation BudCache schedule.

\begin{table}[h]
\centering
\footnotesize
\setlength{\tabcolsep}{2pt}
\caption{Loss objectives at 1,280 updates and NFE 10. Entries are means
$\pm$ sample standard deviations across three fixed training seeds on
the 200 test captions. Bold marks the best mean.}
\label{tab:corrector-loss-objectives}
\begin{tabularx}{\columnwidth}{@{}l*{3}{>{\centering\arraybackslash}X}@{}}
\toprule
Objective & LPIPS$\downarrow$ & SSIM$\uparrow$ & PSNR (dB)$\uparrow$ \\
\midrule
Hidden & $0.1399\pm0.0001$ & $0.8267\pm0.0001$ & $24.758\pm0.008$ \\
Velocity & $0.1365\pm0.0053$ & $0.8310\pm0.0034$ & $24.889\pm0.147$ \\
Joint (Ours) & $\mathbf{0.1358\pm0.0003}$ & $\mathbf{0.8352\pm0.0002}$ & $\mathbf{24.904\pm0.010}$ \\
\bottomrule
\end{tabularx}
\end{table}

Table~\ref{tab:corrector-loss-objectives} shows that joint training (Ours setting) achieves the best mean across all three metrics (0.1358 LPIPS, 0.8352 SSIM, and 24.904\,dB PSNR). It improves all three fidelity metrics over hidden-only training, with paired 95\% intervals excluding zero. Relative to
velocity-only training, joint training improves SSIM by $0.0042$ (95\% CI
$[0.00303,0.00539]$) while also attaining better mean LPIPS and PSNR
($[-0.00203,0.00079]$ for LPIPS and $[-0.03054,0.06202]$\,dB for PSNR across
2,000 paired prompt-bootstrap resamples after averaging the three seeds
within each caption). 
In addition, joint training exhibits substantially smaller dispersion across the three fixed seeds: the standard deviation of seed-mean LPIPS is 0.00026,
versus 0.00531 for velocity-only training. Combining hidden-state and
output-velocity supervision therefore delivers both the highest fidelity and the most consistent optimization.

\subsection{Plug-and-Play Correction for Other Caches}
\label{app:corrector-transfer}
We attach the trained OnCache corrector to BudCache and DiCache, keeping its OnCache-trained weights and $\eta=1$ fixed. BudCache retains its ten-evaluation schedule; DiCache retains its native probe, threshold, and feature extrapolation. Correction adjusts the reused
image-token hidden state before the output head. All results use the same
200 captions at $512^2$ resolution, 30 sampling steps, seed 42, and full 30-step references.

\begin{table}[h]
\centering
\footnotesize
\setlength{\tabcolsep}{2pt}
\caption{Frozen corrector transfer on the 200 captions. NFE counts full evaluations and excludes DiCache's one-block probes.}
\vspace{-6pt}
\label{tab:corrector-transfer}
\begin{tabularx}{\columnwidth}{@{}Xcccc@{}}
\toprule
Method & NFE & LPIPS$\downarrow$ & SSIM$\uparrow$ & PSNR$\uparrow$ \\
\midrule
BudCache & 10.000 & 0.1452 & 0.8255 & 24.573 \\
\quad + corrector & 10.000 & 0.1286 & 0.8480 & 25.198 \\
\midrule
DiCache & 10.005 & 0.1904 & 0.7963 & 22.712 \\
\quad + corrector & 10.000 & 0.1859 & 0.8074 & 23.242 \\
\bottomrule
\end{tabularx}
\vspace{-10pt}
\end{table}

\begin{figure}[!ht]
    \centering
    \includegraphics[width=\columnwidth]{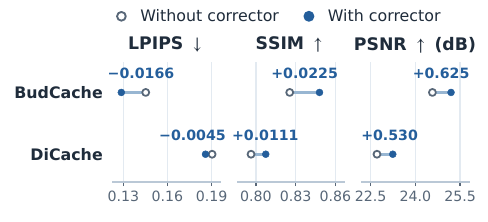}
    \vspace{-16pt}
    \caption{\textbf{Plug-and-play corrector gains.} Paired markers show
    mean fidelity before and after adding the same frozen corrector.}
    \label{fig:corrector-transfer}
    \vspace{-3pt}
\end{figure}

Correction improves all three fidelity metrics for both BudCache and DiCache
(Table~\ref{tab:corrector-transfer}). On BudCache, it reduces LPIPS by
$0.01667$ (95\% CI $[-0.01957,-0.01405]$) while also improving SSIM
and PSNR. On DiCache, correction improves all three metrics, reducing LPIPS by $0.00450$ and raising SSIM and PSNR by $0.01108$ and $0.530$\,dB, respectively, with all paired 95\% intervals excluding zero.
Figure~\ref{fig:corrector-transfer} visualizes how the same corrector lowers LPIPS and raises SSIM and PSNR for both caches at nearly unchanged NFE.

Correction changes 118 of 200 DiCache action sequences. Holding each
original sequence fixed yields a larger LPIPS reduction of $0.00915$
($[-0.01365,-0.00474]$), together with SSIM and PSNR gains whose intervals
exclude zero. These results show that correction transfers across cache mechanisms, with its gain interacting with dynamic scheduling. Intervals use 2,000 prompt-bootstrap resamples.

\section{Model Size and Computational Cost}
\label{app:module-cost}
Table~\ref{tab:module-cost} reports module sizes and online generation times.
Here we detail the offline collection and optimization costs for the FLUX
and Wan policies and correctors.

\subsection{Offline Collection and Optimization}
\label{app:policy-cost}
Table~\ref{tab:inherited-policy-cost} reports offline preparation costs,
with stage times summing to 224.35\,min for FLUX and 306.92\,min for Wan.

\begin{table}[h]
\centering
\small
\setlength{\tabcolsep}{4pt}
\caption{Offline collection and optimization time (minutes) on a single NVIDIA A100 GPU. Totals sum recorded stage durations.}
\vspace{-5pt}
\label{tab:inherited-policy-cost}
\label{tab:corrector-cost}
\label{tab:wan-training-cost}
\begin{tabularx}{0.78\columnwidth}{@{}Xcc@{}}
\toprule
Stage & FLUX & Wan \\
\midrule
Policy data collection & 56.36 & 61.35 \\
Policy fitting and optimization & 53.45 & 95.28 \\
Policy total & 109.81 & 156.63 \\
\midrule
Corrector teacher trajectory collection & 37.71 & 54.11 \\
Corrector fitting & 76.83 & 96.18 \\
Corrector total & 114.54 & 150.29 \\
\bottomrule
\end{tabularx}
\end{table}

\textbf{Policy collection and optimization.}
Collection includes trajectory generation and terminal scoring;
optimization covers policy fitting and offset calibration.

\textbf{Corrector collection and optimization.}
Offline teacher queries provide same-state supervision
(Appendix~\ref{app:corrector-training}).
Collection time spans the first to last trajectory completion;
fitting time sums epoch durations across training.

\subsection{Comparison with Training-Based Caching}
\label{app:training-cache-comparison}
We compare OnCache against ReCache~\citep{DBLP:journals/corr/abs-2606-06060},
a training-based caching baseline whose released 50-step, 13-evaluation
schedules pair with first- and second-order TaylorSeer (T1/T2). All
methods use the same FP16 FLUX backbone, seed~42, guidance~3.5, native
50-step sampler, and Full50 references on one A100 GPU, evaluated on 80
COCO captions disjoint from all teacher-trajectory and fitting sets.

\vspace{-8pt}
\paragraph{Setup.}
For this 50-step setting, we construct OnCache from a 13-evaluation DP
prior on 64 $512^2$ teacher-trajectory captions (using the same paired
policy-fitting pipeline and a 40-epoch corrector fit with cosine learning rate from $10^{-4}$ to $10^{-5}$), with a
global offset calibrated toward a mean budget of 13~NFE. All weights, normalizers, offsets, and $\eta=1$ are frozen before the 80-caption evaluation.

\begin{table}[t]
\centering
\footnotesize
\setlength{\tabcolsep}{2pt}
\caption{50-step comparison on 80 $512^2$ captions. Time includes VAE decoding. Bold marks
the best score in fidelity column.}
\vspace{-5pt}
\label{tab:training-cache-comparison}
\begin{tabularx}{\columnwidth}{@{}Xcccccc@{}}
\toprule
Method & NFE & LPIPS$\downarrow$ & SSIM$\uparrow$ & PSNR$\uparrow$ & Time (s)$\downarrow$ & Speedup$\uparrow$ \\
\midrule
Full50 & 50.000 & -- & -- & -- & 8.115 & 1.00$\times$ \\
ReCache + T1 & 13.000 & 0.1293 & 0.8478 & 23.882 & 2.742 & 2.96$\times$ \\
ReCache + T2 & 13.000 & 0.1364 & 0.8442 & 23.494 & 2.958 & 2.74$\times$ \\
\midrule
OnCache (w/o) & 13.238 & 0.1362 & 0.8426 & 23.561 & 2.394 & 3.39$\times$ \\
OnCache (w) & 13.050 & \textbf{0.1272} & \textbf{0.8495} & \textbf{24.026} & 2.437 & 3.33$\times$ \\
\bottomrule
\end{tabularx}
\end{table}

\vspace{-8pt}
\paragraph{Inference speed and fidelity.}
As shown in Table~\ref{tab:training-cache-comparison}, OnCache (w) achieves the best generation quality across all three fidelity metrics (0.1272 LPIPS, 0.8495 SSIM, and 24.026\,dB PSNR), outperforming both ReCache + TaylorSeer Order 1 (T1) and ReCache + TaylorSeer Order 2 (T2) while running markedly faster. At virtually the same $\sim$13~NFE, OnCache (w/o) and OnCache (w) reach $3.39\times$ and $3.33\times$ speedups over Full50, compared with $2.96\times$ and $2.74\times$ for ReCache + T1/T2. This large latency gap arises because ReCache relies on online block-wise TaylorSeer feature forecasting across Transformer blocks at every cached step, whereas OnCache reuses the whole-stack residual with a lightweight corrector that adds only 43\,ms per image.

\begin{table}[t]
\centering
\small
\setlength{\tabcolsep}{3pt}
\caption{Offline training cost for the 50-step comparison. ReCache reports H100 GPU hours; OnCache reports A100 GPU hours.}
\vspace{-5pt}
\label{tab:training-cache-cost}
\begin{tabularx}{0.85\columnwidth}{@{}Xcc@{}}
\toprule
Configuration & Device & Hours \\
\midrule
ReCache + T1 (reported) & H100 & 22 \\
ReCache + T2 (reported) & H100 & 23 \\
OnCache (w/o) & A100 & $\approx2.28$ \\
OnCache (w) & A100 & $\approx2.81$ \\
\bottomrule
\end{tabularx}
\vspace{-6pt}
\end{table}

\paragraph{Offline optimization efficiency.}
Table~\ref{tab:training-cache-cost} highlights a decisive advantage in
training efficiency: whereas ReCache requires 22--23 H100 GPU hours for
offline schedule learning, OnCache completes all offline data collection,
policy and corrector optimization in less than 3 A100 GPU hours.

\section{Extended Related Work}
\label{app:related}
Diffusion inference can be accelerated through faster numerical samplers
such as DDIM, PNDM, EDM, DPM-Solver, DPM-Solver++, and
UniPC~\citep{song2021ddim,DBLP:conf/iclr/Liu0LZ22,DBLP:conf/nips/KarrasAAL22,lu2022dpmsolver,DBLP:journals/ijautcomp/LuZBCLZ25,DBLP:conf/nips/ZhaoBR0L23},
or by distilling pretrained models into one- or few-step generators via
progressive, consistency, and distribution-matching
distillation~\citep{salimans2022progressive,DBLP:conf/icml/SongD0S23,DBLP:journals/corr/abs-2310-04378,DBLP:conf/iclr/LuS25,DBLP:conf/cvpr/YinG0SDFP24,DBLP:conf/nips/YinGP0SDF24}.
Orthogonal to these approaches, feature caching exploits temporal
redundancy within the denoising steps by reusing intermediate representations. Below, we expand on the caching, correction, and sequential policy learning comparisons summarized in Sec.~\ref{sec:related}.

\subsection{Reuse Granularity and Online Decisions}
\label{app:related-reuse}
DeepCache enables training-free reuse of high-level U-Net features while updating the shallow branch~\citep{DBLP:conf/cvpr/MaFW24}. Block Caching adds block-level reuse and learned scale-and-shift adjustments~\citep{DBLP:conf/cvpr/WimbauerWSDHHSZ24}; FORA reuses attention and MLP intermediates~\citep{DBLP:journals/corr/abs-2407-01425}. Learning-to-Cache (L2C) differentiably learns a time-dependent, input-invariant layer router, producing a static deployment graph~\citep{DBLP:conf/nips/MaFMW24}. ToCa recomputes important tokens while reusing the others, varying caching ratios across layer types and depths~\citep{DBLP:conf/iclr/ZouLLHZ25}. DuCa alternates whole-feature reuse with conservative token refresh using random token selection~\citep{DBLP:journals/tip/ZouZZGXSHHZ26}. These designs vary the granularity of reused computation. OnCache reuses the complete Transformer block-stack residual and schedules full-stack evaluations.

Online adaptation supplies feedback from the trajectory being generated.
TeaCache accumulates rescaled differences in timestep-modulated inputs to trigger refresh~\citep{liu2025teacache}, SeaCache filters feature changes through spectral evolution to separate content-relevant variation from
noise~\citep{chung2026seacache}, DiCache probes shallow layers for error
estimation and trajectory alignment~\citep{bu2026dicache}. These methods
adapt refresh decisions to the running trajectory and allow NFE to vary across inputs. OnCache likewise conditions decisions on the current state via an offline-fitted policy.

\subsection{Prediction, Correction, and Scheduling}
\vspace{-2pt}
\label{app:related-prediction}
TaylorSeer extrapolates feature history through Taylor expansions~\citep{liu2025taylorseer}, Spectrum fits temporal features in a Chebyshev basis through regularized regression~\citep{han2026spectrum}. Both go beyond direct cache copying. MagCache controls reuse using calibrated residual magnitude-ratio profiles~\citep{DBLP:conf/nips/MaWWZT25}. ERTACache combines residual error rectification with trajectory-aware timestep adjustment~\citep{peng2026ertacache}. EpaCache measures final latent deviation after an isolated cache action followed by full evaluations, then aggregates results across calibration prompts into timestep-dependent refresh tolerances~\citep{DBLP:journals/corr/abs-2608-29264}. OnCache trains a corrector to predict state-dependent adjustments.

DPCache formulates denoising as path planning with dynamic programming over calibrated interval costs~\citep{cui2026dpcache}; BudCache uses simulated annealing and deterministic hill climbing under an explicit budget~\citep{lei2026budcache}. Both use a prescribed NFE. ReCache learns budget-conditioned schedules with policy gradients and terminal fidelity and quality rewards, deploying a shared schedule per budget~\citep{DBLP:journals/corr/abs-2606-06060}. BAG distills binary decisions from searched prompt-budget schedules into online gates conditioned on budget state and local feedback, enforcing an exact per-image NFE~\citep{DBLP:journals/corr/abs-2608-09231}. OnCache fits fidelity gains and actual NFE differences from paired continuations, then refines a policy using new states it visits.

\subsection{Connections to Sequential Policy Learning}
\vspace{-2pt}
\label{app:related-sequential}
An online decision changes the states at which later decisions are made.
DAgger addresses this dependence through iterative data aggregation under
the learned policy~\citep{DBLP:journals/jmlr/RossGB11}; AGGREVATE incorporates
cost-to-go information into interactive cost-sensitive
learning~\citep{DBLP:journals/corr/RossB14}. This perspective motivates
querying alternative actions at states reached by the collection
policy. As the policy changes, both the visited states and their
continuation-dependent values change, motivating the refreshed supervision. 
Outcome comparisons also connect paired-continuation supervision to
preference learning~\citep{rafailov2023dpo}. Here, paired continuations provide
regression targets for fidelity gains and total NFE changes, linking
state-dependent decisions to final generation quality and computation.
\clearpage
\onecolumn
\setcounter{figure}{0}
\renewcommand{\thefigure}{S\arabic{figure}}
\section{Qualitative Comparisons}
\label{app:qualitative}

\begin{figure}[htbp]
    \centering
    \includegraphics[width=\linewidth]{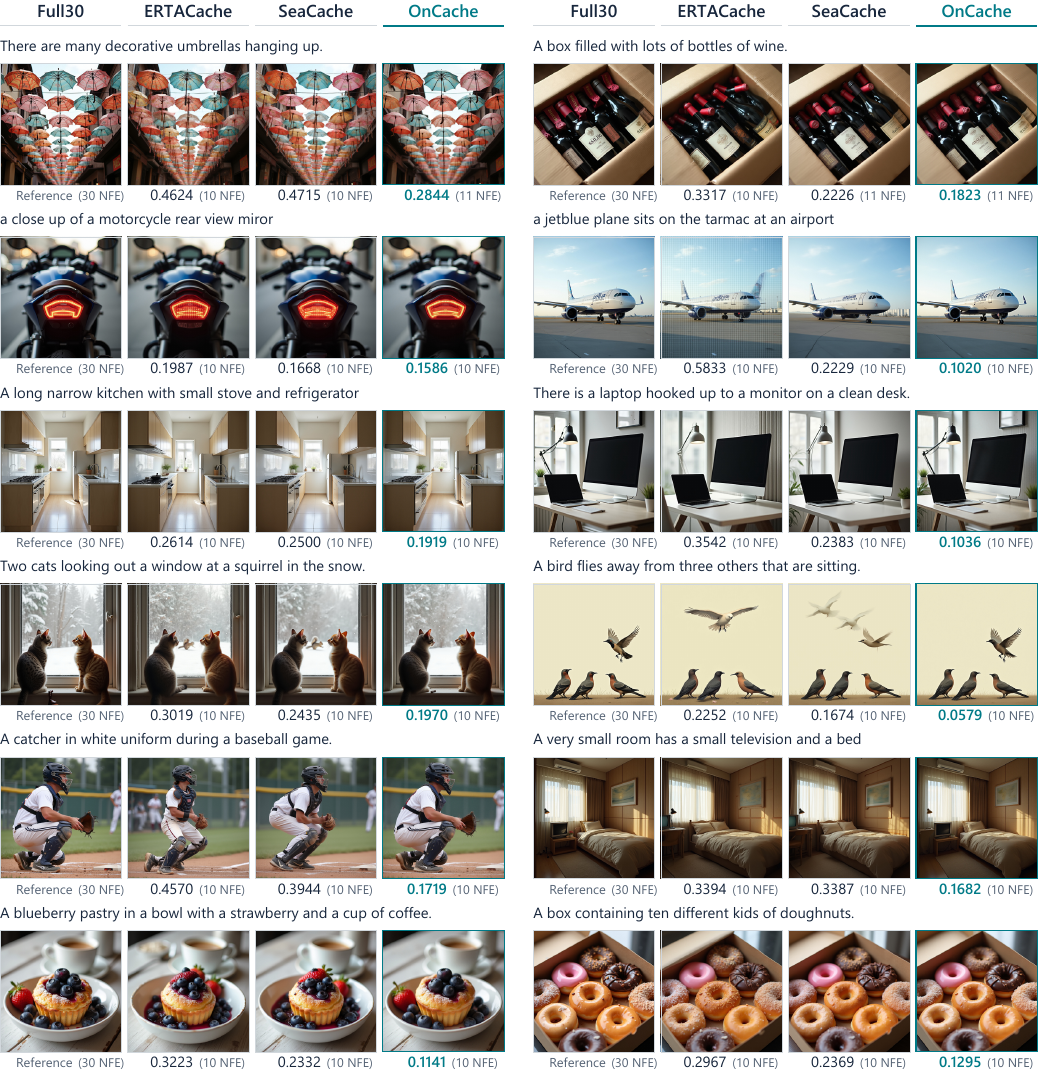}
    \caption{\textbf{Selected comparisons at $1024^2$ resolution.}
    The OnCache column (without corrector) has lower LPIPS than ERTACache and SeaCache on all twelve prompts.
    Values give full-image LPIPS to Full30 ($\downarrow$), with actual NFE in parentheses.}
    \label{fig:qualitative-native-a}
\end{figure}
\clearpage

\begin{figure}[htbp]
    \centering
    \includegraphics[width=\linewidth]{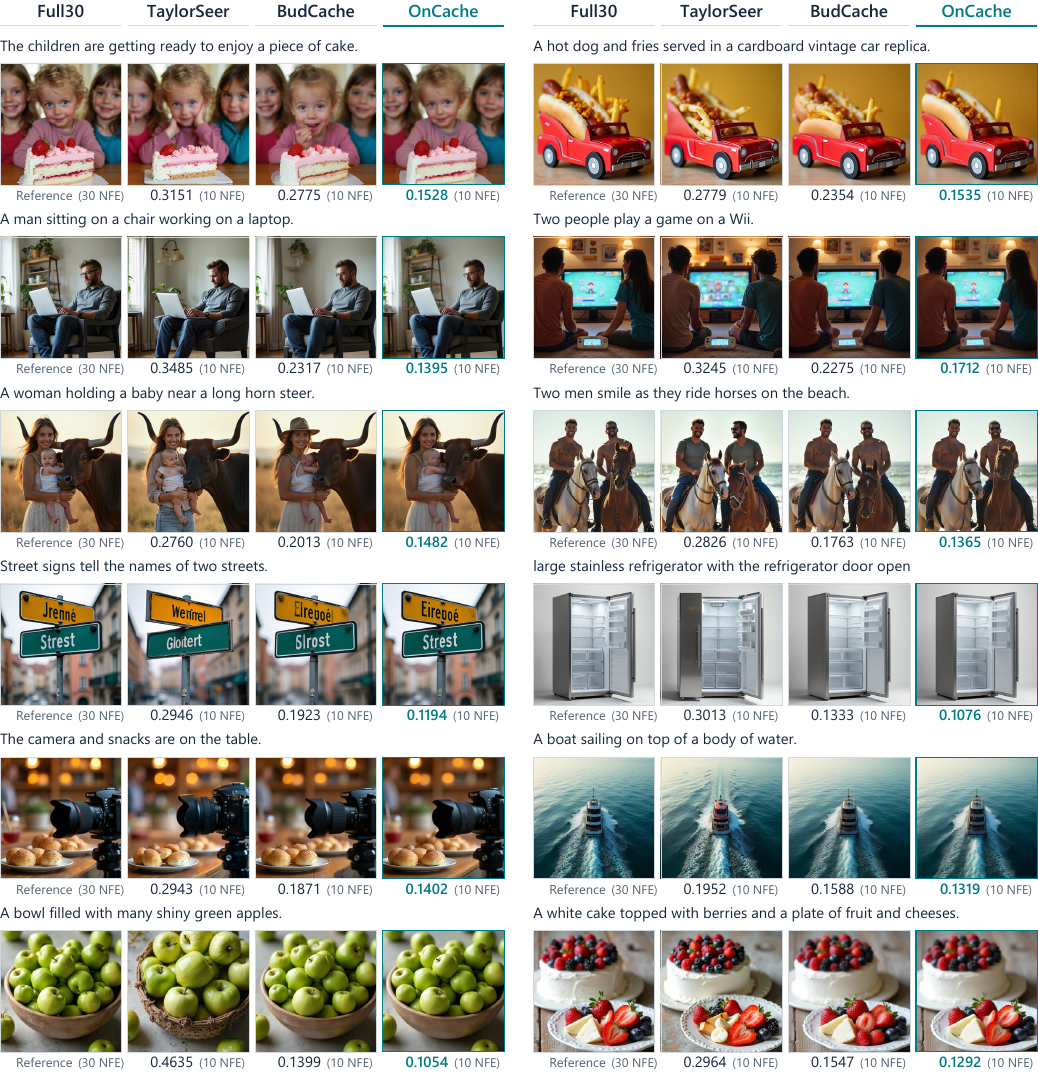}
    \caption{\textbf{Comparisons at $512^2$ resolution.} The OnCache column (without corrector) has lower LPIPS than TaylorSeer and BudCache on all twelve prompts at an NFE of ten per accelerated output. Values give full-image LPIPS to Full30 ($\downarrow$), with actual NFE in parentheses.}
    \label{fig:qualitative-native-512}
\end{figure}
\clearpage

\begin{figure}[p]
    \centering
    \includegraphics[width=\linewidth,height=.88\textheight,keepaspectratio]{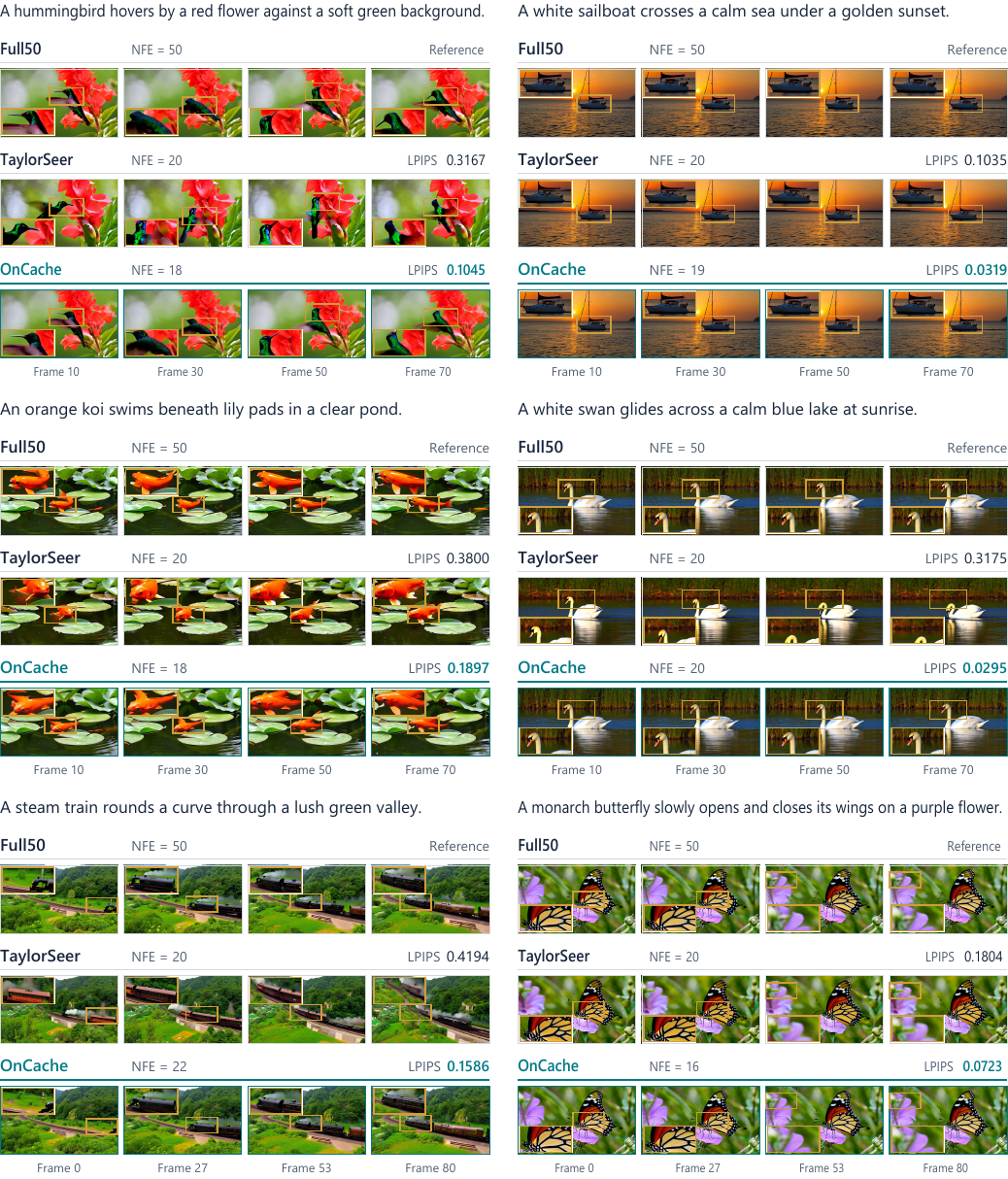}
    \caption{\textbf{Selected Wan2.1-T2V-1.3B video comparisons.}
    Each example shows four evenly spaced frames from paired videos
    (seed 0, $832\times480$, 81 frames, 50 sampling steps). Values report actual NFE and mean LPIPS over all decoded
    frames against Full50. Gold boxes and insets mark matching regions; the
    train and butterfly show additional frames from the videos in the lower panel of Fig.~\ref{fig:qualitative-native}.}
    \label{fig:qualitative-wan-appendix}
\end{figure}
\clearpage

%% file: ref.bib
@inproceedings{rombach2022latent,
  author = {Robin Rombach and Andreas Blattmann and Dominik Lorenz and Patrick Esser and Bj{\"o}rn Ommer},
  title = {High-Resolution Image Synthesis with Latent Diffusion Models},
  booktitle = {Proceedings of the {IEEE/CVF} Conference on Computer Vision and Pattern Recognition},
  pages = {10684--10695},
  year = {2022},
  url = {https://arxiv.org/abs/2112.10752}
}

@inproceedings{song2021ddim,
  author = {Jiaming Song and Chenlin Meng and Stefano Ermon},
  title = {Denoising Diffusion Implicit Models},
  booktitle = {International Conference on Learning Representations},
  year = {2021},
  url = {https://arxiv.org/abs/2010.02502}
}

@inproceedings{salimans2022progressive,
  author = {Tim Salimans and Jonathan Ho},
  title = {Progressive Distillation for Fast Sampling of Diffusion Models},
  booktitle = {International Conference on Learning Representations},
  year = {2022},
  url = {https://arxiv.org/abs/2202.00512}
}

@inproceedings{lu2022dpmsolver,
  author = {Cheng Lu and Yuhao Zhou and Fan Bao and Jianfei Chen and Chongxuan Li and Jun Zhu},
  title = {{DPM-Solver}: A Fast {ODE} Solver for Diffusion Probabilistic Model Sampling in Around 10 Steps},
  booktitle = {Advances in Neural Information Processing Systems},
  volume = {35},
  pages = {5775--5787},
  year = {2022},
  url = {https://arxiv.org/abs/2206.00927}
}

@inproceedings{DBLP:conf/cvpr/HuangHYZS0Z0JCW24,
  author       = {Ziqi Huang and
                  Yinan He and
                  Jiashuo Yu and
                  Fan Zhang and
                  Chenyang Si and
                  Yuming Jiang and
                  Yuanhan Zhang and
                  Tianxing Wu and
                  Qingyang Jin and
                  Nattapol Chanpaisit and
                  Yaohui Wang and
                  Xinyuan Chen and
                  Limin Wang and
                  Dahua Lin and
                  Yu Qiao and
                  Ziwei Liu},
  title        = {VBench: Comprehensive Benchmark Suite for Video Generative Models},
  booktitle    = {Proceedings of the {IEEE/CVF} Conference on Computer Vision and Pattern Recognition},
  pages        = {21807--21818},
  publisher    = {{IEEE}},
  year         = {2024}
}

@inproceedings{DBLP:conf/cvpr/MaFW24,
  author       = {Xinyin Ma and
                  Gongfan Fang and
                  Xinchao Wang},
  title        = {{DeepCache}: Accelerating Diffusion Models for Free},
  booktitle    = {Proceedings of the {IEEE/CVF} Conference on Computer Vision and Pattern Recognition},
  pages        = {15762--15772},
  publisher    = {{IEEE}},
  year         = {2024}
}

@inproceedings{DBLP:conf/cvpr/WimbauerWSDHHSZ24,
  author       = {Felix Wimbauer and
                  Bichen Wu and
                  Edgar Sch{\"{o}}nfeld and
                  Xiaoliang Dai and
                  Ji Hou and
                  Zijian He and
                  Artsiom Sanakoyeu and
                  Peizhao Zhang and
                  Sam S. Tsai and
                  Jonas Kohler and
                  Christian Rupprecht and
                  Daniel Cremers and
                  Peter Vajda and
                  Jialiang Wang},
  title        = {Cache Me if You Can: Accelerating Diffusion Models through Block Caching},
  booktitle    = {Proceedings of the {IEEE/CVF} Conference on Computer Vision and Pattern Recognition},
  pages        = {6211--6220},
  publisher    = {{IEEE}},
  year         = {2024}
}

@inproceedings{DBLP:conf/cvpr/ZhangIESW18,
  author       = {Richard Zhang and
                  Phillip Isola and
                  Alexei A. Efros and
                  Eli Shechtman and
                  Oliver Wang},
  title        = {The Unreasonable Effectiveness of Deep Features as a Perceptual Metric},
  booktitle    = {Proceedings of the {IEEE} Conference on Computer Vision and Pattern Recognition},
  pages        = {586--595},
  publisher    = {Computer Vision Foundation / {IEEE} Computer Society},
  year         = {2018}
}

@inproceedings{DBLP:conf/eccv/LinMBHPRDZ14,
  author       = {Tsung{-}Yi Lin and
                  Michael Maire and
                  Serge J. Belongie and
                  James Hays and
                  Pietro Perona and
                  Deva Ramanan and
                  Piotr Doll{\'{a}}r and
                  C. Lawrence Zitnick},
  editor       = {David J. Fleet and
                  Tom{\'{a}}s Pajdla and
                  Bernt Schiele and
                  Tinne Tuytelaars},
  title        = {Microsoft {COCO:} Common Objects in Context},
  booktitle    = {European Conference on Computer Vision},
  series       = {Lecture Notes in Computer Science},
  volume       = {8693},
  pages        = {740--755},
  publisher    = {Springer},
  year         = {2014}
}

@inproceedings{DBLP:conf/iccv/PeeblesX23,
  author       = {William Peebles and
                  Saining Xie},
  title        = {Scalable Diffusion Models with Transformers},
  booktitle    = {Proceedings of the {IEEE/CVF} International Conference on Computer Vision},
  pages        = {4172--4182},
  publisher    = {{IEEE}},
  year         = {2023}
}

@inproceedings{DBLP:conf/iclr/YangTZ00XYHZFYZ25,
  author       = {Zhuoyi Yang and
                  Jiayan Teng and
                  Wendi Zheng and
                  Ming Ding and
                  Shiyu Huang and
                  Jiazheng Xu and
                  Yuanming Yang and
                  Wenyi Hong and
                  Xiaohan Zhang and
                  Guanyu Feng and
                  Da Yin and
                  Yuxuan Zhang and
                  Weihan Wang and
                  Yean Cheng and
                  Bin Xu and
                  Xiaotao Gu and
                  Yuxiao Dong and
                  Jie Tang},
  title        = {CogVideoX: Text-to-Video Diffusion Models with An Expert Transformer},
  booktitle    = {International Conference on Learning Representations},
  publisher    = {OpenReview.net},
  year         = {2025}
}

@inproceedings{DBLP:conf/nips/HoJA20,
  author       = {Jonathan Ho and
                  Ajay Jain and
                  Pieter Abbeel},
  editor       = {Hugo Larochelle and
                  Marc'Aurelio Ranzato and
                  Raia Hadsell and
                  Maria{-}Florina Balcan and
                  Hsuan{-}Tien Lin},
  title        = {Denoising Diffusion Probabilistic Models},
  booktitle    = {Advances in Neural Information Processing Systems 33},
  year         = {2020}
}

@article{DBLP:journals/tip/WangBSS04,
  author       = {Zhou Wang and
                  Alan C. Bovik and
                  Hamid R. Sheikh and
                  Eero P. Simoncelli},
  title        = {Image quality assessment: from error visibility to structural similarity},
  journal      = {{IEEE} Trans. Image Process.},
  volume       = {13},
  number       = {4},
  pages        = {600--612},
  year         = {2004}
}

@misc{fluxdevmodelcard,
  author = {{Black Forest Labs}},
  title = {{FLUX.1} [dev]},
  year = {2024},
  howpublished = {\url{https://huggingface.co/black-forest-labs/FLUX.1-dev}},
  note = {Model card; accessed November 2, 2026}
}

@inproceedings{liu2025teacache,
  author    = {Feng Liu and Shiwei Zhang and Xiaofeng Wang and Yujie Wei and Haonan Qiu and Yuzhong Zhao and Yingya Zhang and Qixiang Ye and Fang Wan},
  title     = {Timestep Embedding Tells: It's Time to Cache for Video Diffusion Model},
  booktitle = {Proceedings of the {IEEE/CVF} Conference on Computer Vision and Pattern Recognition},
  pages     = {7353--7363},
  year      = {2025},
  doi       = {10.1109/CVPR52734.2025.00689},
  url       = {https://doi.org/10.1109/CVPR52734.2025.00689},
  bibsource = {dblp computer science bibliography, https://dblp.org}
}

@inproceedings{chung2026seacache,
  author    = {Jiwoo Chung and Sangeek Hyun and MinKyu Lee and Byeongju Han and Geonho Cha and Dongyoon Wee and Youngjun Hong and Jae-Pil Heo},
  title     = {{SeaCache}: Spectral-Evolution-Aware Cache for Accelerating Diffusion Models},
  booktitle = {Proceedings of the {IEEE/CVF} Conference on Computer Vision and Pattern Recognition},
  pages     = {14283--14294},
  year      = {2026},
  url       = {https://openaccess.thecvf.com/content/CVPR2026/html/Chung_SeaCache_Spectral-Evolution-Aware_Cache_for_Accelerating_Diffusion_Models_CVPR_2026_paper.html},
  bibsource = {dblp computer science bibliography, https://dblp.org}
}

@inproceedings{han2026spectrum,
  author    = {Han, Jiaqi and Shi, Juntong and Li, Puheng and Ye, Haotian and Guo, Qiushan and Ermon, Stefano},
  title     = {Adaptive Spectral Feature Forecasting for Diffusion Sampling Acceleration},
  booktitle = {Proceedings of the {IEEE/CVF} Conference on Computer Vision and Pattern Recognition},
  year      = {2026},
  pages     = {43320--43330},
  url       = {https://openaccess.thecvf.com/content/CVPR2026/html/Han_Adaptive_Spectral_Feature_Forecasting_for_Diffusion_Sampling_Acceleration_CVPR_2026_paper.html},
  bibsource = {dblp computer science bibliography, https://dblp.org}
}

@inproceedings{cui2026dpcache,
  author    = {Cui, Bowen and Wang, Yuanbin and Xu, Huajiang and Chen, Biaolong and Zhang, Aixi and Jiang, Hao and Jin, Zhengzheng and Liu, Xu and Huang, Pipei},
  title     = {Denoising as Path Planning: Training-Free Acceleration of Diffusion Models with {DPCache}},
  booktitle = {Proceedings of the {IEEE/CVF} Conference on Computer Vision and Pattern Recognition},
  year      = {2026},
  pages     = {43632--43642},
  url       = {https://openaccess.thecvf.com/content/CVPR2026/html/Cui_Denoising_as_Path_Planning_Training-Free_Acceleration_of_Diffusion_Models_with_CVPR_2026_paper.html},
  bibsource = {dblp computer science bibliography, https://dblp.org}
}

@inproceedings{liu2025taylorseer,
  author       = {Jiacheng Liu and
                  Chang Zou and
                  Yuanhuiyi Lyu and
                  Junjie Chen and
                  Linfeng Zhang},
  title        = {From Reusing to Forecasting: Accelerating Diffusion Models With {TaylorSeers}},
  booktitle    = {Proceedings of the {IEEE/CVF} International Conference on Computer Vision},
  pages        = {15853--15863},
  publisher    = {{IEEE}},
  year         = {2025},
  url          = {https://doi.org/10.1109/ICCV51701.2025.01471},
  doi          = {10.1109/ICCV51701.2025.01471},
  bibsource    = {dblp computer science bibliography, https://dblp.org}
}

@inproceedings{peng2026ertacache,
  author    = {Peng, Xurui and Yan, Chenqian and Liu, Hong and Ma, Rui and Chen, Fangmin and Wang, Xing and Wu, Zhihua and Liu, Songwei and Lin, Mingbao},
  title     = {{ERTACache}: Error Rectification and Timesteps Adjustment for Efficient Diffusion},
  booktitle = {International Conference on Learning Representations},
  editor    = {C. Vondrick and B. Hariharan and C. Raffel and L. Pinto and D. Yang and A. Faust},
  volume    = {2026},
  year      = {2026},
  pages     = {82104--82120},
  url       = {https://proceedings.iclr.cc/paper_files/paper/2026/file/84d395725a9b40cb4a49d84478ac24c7-Paper-Conference.pdf},
  bibsource = {dblp computer science bibliography, https://dblp.org; publisher proceedings for additional fields}
}

@inproceedings{bu2026dicache,
  author    = {Jiazi Bu and Pengyang Ling and Yujie Zhou and Yibin Wang and Yuhang Zang and Dahua Lin and Jiaqi Wang},
  title     = {{DiCache}: Let Diffusion Model Determine Its Own Cache},
  booktitle = {International Conference on Learning Representations},
  year      = {2026},
  url       = {https://dblp.org/rec/conf/iclr/BuLZWZLW26},
  bibsource = {dblp computer science bibliography, https://dblp.org}
}

@inproceedings{lei2026budcache,
  author    = {Lei, Mingkun and Zhao, Tong and Yuan, Liangyu and Zhang, Chi},
  title     = {Budget-Constrained Step-Level Diffusion Caching},
  booktitle = {Proceedings of the 43rd International Conference on Machine Learning},
  editor    = {Zhang, Tong and Dudik, Miroslav and Jaggi, Martin and Agarwal, Alekh and Li, Sharon and Schuurmans, Dale and Zhu, Jerry and Berkenkamp, Felix and Dong, Hanze and Bietti, Alberto},
  volume    = {306},
  series    = {Proceedings of Machine Learning Research},
  publisher = {PMLR},
  month     = {06--11 Jul},
  year      = {2026},
  pages     = {65536--65553},
  url       = {https://proceedings.mlr.press/v306/lei26j.html},
  bibsource = {Official PMLR proceedings; DBLP association is the CoRR preprint only}
}

@inproceedings{xu2023imagereward,
  author    = {Xu, Jiazheng and Liu, Xiao and Wu, Yuchen and Tong, Yuxuan and Li, Qinkai and Ding, Ming and Tang, Jie and Dong, Yuxiao},
  title     = {{ImageReward}: Learning and Evaluating Human Preferences for Text-to-Image Generation},
  booktitle = {Advances in Neural Information Processing Systems},
  editor    = {A. Oh and T. Naumann and A. Globerson and K. Saenko and M. Hardt and S. Levine},
  volume    = {36},
  pages     = {15903--15935},
  publisher = {Curran Associates, Inc.},
  year      = {2023},
  doi       = {10.52202/075280-0700},
  url       = {https://proceedings.neurips.cc/paper_files/paper/2023/hash/33646ef0ed554145eab65f6250fab0c9-Abstract-Conference.html}
}

@inproceedings{DBLP:conf/nips/MaFMW24,
  author = {Xinyin Ma and Gongfan Fang and Michael Bi Mi and Xinchao Wang},
  title = {{Learning-to-Cache}: Accelerating Diffusion Transformer via Layer Caching},
  booktitle = {Advances in Neural Information Processing Systems},
  year = {2024},
  url = {https://arxiv.org/abs/2406.01733},
  bibsource = {dblp computer science bibliography, https://dblp.org}
}

@article{DBLP:journals/corr/abs-2407-01425,
  author = {Pratheba Selvaraju and Tianyu Ding and Tianyi Chen and Ilya Zharkov and Luming Liang},
  title = {{FORA}: Fast-Forward Caching in Diffusion Transformer Acceleration},
  journal = {CoRR},
  volume = {abs/2407.01425},
  year = {2024},
  doi = {10.48550/arXiv.2407.01425},
  url = {https://arxiv.org/abs/2407.01425},
  bibsource = {dblp computer science bibliography, https://dblp.org}
}

@inproceedings{DBLP:conf/iclr/ZouLLHZ25,
  author = {Chang Zou and Xuyang Liu and Ting Liu and Siteng Huang and Linfeng Zhang},
  title = {Accelerating Diffusion Transformers with Token-wise Feature Caching},
  booktitle = {International Conference on Learning Representations},
  publisher = {OpenReview.net},
  year = {2025},
  url = {https://openreview.net/forum?id=yYZbZGo4ei},
  bibsource = {dblp computer science bibliography, https://dblp.org}
}

@article{DBLP:journals/tip/ZouZZGXSHHZ26,
  author = {Chang Zou and Shikang Zheng and Evelyn Zhang and Runlin Guo and Haohang Xu and Zhengyi Shi and Conghui He and Xuming Hu and Linfeng Zhang},
  title = {Rethinking Token-Wise Feature Caching: Accelerating Diffusion Transformers With Dual Feature Caching},
  journal = {{IEEE} Trans. Image Process.},
  volume = {35},
  pages = {6211--6220},
  year = {2026},
  doi = {10.1109/TIP.2026.3698363},
  url = {https://doi.org/10.1109/TIP.2026.3698363},
  bibsource = {dblp computer science bibliography, https://dblp.org}
}

@inproceedings{DBLP:conf/nips/MaWWZT25,
  author = {Zehong Ma and Longhui Wei and Feng Wang and Shiliang Zhang and Qi Tian},
  title = {{MagCache}: Fast Video Generation with Magnitude-Aware Cache},
  booktitle = {Advances in Neural Information Processing Systems},
  year = {2025},
  url = {https://arxiv.org/abs/2506.09045},
  bibsource = {dblp computer science bibliography, https://dblp.org}
}

@inproceedings{DBLP:journals/jmlr/RossGB11,
  author = {St{\'{e}}phane Ross and Geoffrey J. Gordon and Drew Bagnell},
  title = {A Reduction of Imitation Learning and Structured Prediction to No-Regret Online Learning},
  booktitle = {Proceedings of the Fourteenth International Conference on Artificial Intelligence and Statistics, {AISTATS} 2011},
  series = {{JMLR} Proceedings},
  volume = {15},
  pages = {627--635},
  publisher = {JMLR.org},
  year = {2011},
  url = {https://proceedings.mlr.press/v15/ross11a.html},
  bibsource = {dblp computer science bibliography, https://dblp.org}
}

@article{DBLP:journals/corr/RossB14,
  author = {St{\'{e}}phane Ross and J. Andrew Bagnell},
  title = {Reinforcement and Imitation Learning via Interactive No-Regret Learning},
  journal = {CoRR},
  volume = {abs/1406.5979},
  year = {2014},
  url = {https://arxiv.org/abs/1406.5979},
  bibsource = {dblp computer science bibliography, https://dblp.org}
}

@article{DBLP:journals/corr/abs-2608-09231,
  author = {Tong Zhao and Mingkun Lei and Yucheng Han and Chi Zhang},
  title = {{BAG}: Budget-Aware Gating for Diffusion Caching},
  journal = {CoRR},
  volume = {abs/2608.09231},
  year = {2026},
  url = {https://doi.org/10.48550/arXiv.2608.09231},
  doi = {10.48550/ARXIV.2608.09231},
  eprinttype = {arXiv},
  eprint = {2608.09231},
  bibsource = {dblp computer science bibliography, https://dblp.org}
}

@article{DBLP:journals/corr/abs-2608-29264,
  author = {Yuhan Liu and Zongwei Hong and Jinglun Li and Linze Li and Shen Zhang and Yao Tang},
  title = {{EpaCache}: Error-Propagation-Aware Caching for Accelerating Diffusion-Based Visual Generation},
  journal = {CoRR},
  volume = {abs/2608.29264},
  year = {2026},
  url = {https://doi.org/10.48550/arXiv.2608.29264},
  doi = {10.48550/ARXIV.2608.29264},
  eprinttype = {arXiv},
  eprint = {2608.29264},
  bibsource = {dblp computer science bibliography, https://dblp.org}
}

@misc{rafailov2023dpo,
  author = {Rafael Rafailov and Archit Sharma and Eric Mitchell and Stefano Ermon and Christopher D. Manning and Chelsea Finn},
  title = {Direct Preference Optimization: Your Language Model is Secretly a Reward Model},
  year = {2023},
  eprint = {2305.18290},
  archivePrefix = {arXiv},
  primaryClass = {cs.LG},
  url = {https://arxiv.org/abs/2305.18290}
}

@article{DBLP:journals/corr/abs-2606-06060,
  author = {Mishan Aliev and Eva Neudachina and Ilya Bykov and Aleksandr Oganov and Kirill Struminsky and Aibek Alanov and Denis Rakitin},
  title = {{ReCache}: Learning Budget-Aware Caching Schedules for Diffusion Models via {REINFORCE}},
  journal = {CoRR},
  volume = {abs/2606.06060},
  year = {2026},
  url = {https://doi.org/10.48550/arXiv.2606.06060},
  doi = {10.48550/ARXIV.2606.06060},
  eprinttype = {arXiv},
  eprint = {2606.06060},
  bibsource = {dblp computer science bibliography, https://dblp.org}
}

@inproceedings{DBLP:conf/iclr/Liu0LZ22,
  author = {Luping Liu and Yi Ren and Zhijie Lin and Zhou Zhao},
  title = {Pseudo Numerical Methods for Diffusion Models on Manifolds},
  booktitle = {International Conference on Learning Representations},
  year = {2022},
  url = {https://openreview.net/forum?id=PlKWVd2yBkY},
  bibsource = {dblp computer science bibliography, https://dblp.org}
}

@inproceedings{DBLP:conf/nips/KarrasAAL22,
  author = {Tero Karras and Miika Aittala and Timo Aila and Samuli Laine},
  title = {Elucidating the Design Space of Diffusion-Based Generative Models},
  booktitle = {Advances in Neural Information Processing Systems},
  volume = {35},
  year = {2022},
  url = {https://papers.nips.cc/paper_files/paper/2022/hash/a98846e9d9cc01cfb87eb694d946ce6b-Abstract-Conference.html},
  bibsource = {dblp computer science bibliography, https://dblp.org}
}

@article{DBLP:journals/ijautcomp/LuZBCLZ25,
  author = {Cheng Lu and Yuhao Zhou and Fan Bao and Jianfei Chen and Chongxuan Li and Jun Zhu},
  title = {{DPM-Solver++}: Fast Solver for Guided Sampling of Diffusion Probabilistic Models},
  journal = {Mach. Intell. Res.},
  volume = {22},
  number = {4},
  pages = {730--751},
  year = {2025},
  url = {https://doi.org/10.1007/s11633-025-1562-4},
  doi = {10.1007/S11633-025-1562-4},
  bibsource = {dblp computer science bibliography, https://dblp.org}
}

@inproceedings{DBLP:conf/nips/ZhaoBR0L23,
  author = {Wenliang Zhao and Lujia Bai and Yongming Rao and Jie Zhou and Jiwen Lu},
  title = {{UniPC}: A Unified Predictor-Corrector Framework for Fast Sampling of Diffusion Models},
  booktitle = {Advances in Neural Information Processing Systems},
  volume = {36},
  year = {2023},
  url = {https://papers.nips.cc/paper_files/paper/2023/hash/9c2aa1e456ea543997f6927295196381-Abstract-Conference.html},
  bibsource = {dblp computer science bibliography, https://dblp.org}
}

@inproceedings{DBLP:conf/icml/SongD0S23,
  author = {Yang Song and Prafulla Dhariwal and Mark Chen and Ilya Sutskever},
  title = {Consistency Models},
  booktitle = {International Conference on Machine Learning},
  series = {Proceedings of Machine Learning Research},
  volume = {202},
  pages = {32211--32252},
  publisher = {PMLR},
  year = {2023},
  url = {https://proceedings.mlr.press/v202/song23a.html},
  bibsource = {dblp computer science bibliography, https://dblp.org}
}

@article{DBLP:journals/corr/abs-2310-04378,
  author = {Simian Luo and Yiqin Tan and Longbo Huang and Jian Li and Hang Zhao},
  title = {Latent Consistency Models: Synthesizing High-Resolution Images with Few-Step Inference},
  journal = {CoRR},
  volume = {abs/2310.04378},
  year = {2023},
  url = {https://doi.org/10.48550/arXiv.2310.04378},
  doi = {10.48550/ARXIV.2310.04378},
  eprinttype = {arXiv},
  eprint = {2310.04378},
  bibsource = {dblp computer science bibliography, https://dblp.org}
}

@inproceedings{DBLP:conf/iclr/LuS25,
  author = {Cheng Lu and Yang Song},
  title = {Simplifying, Stabilizing and Scaling Continuous-time Consistency Models},
  booktitle = {International Conference on Learning Representations},
  year = {2025},
  url = {https://openreview.net/forum?id=LyJi5ugyJx},
  bibsource = {dblp computer science bibliography, https://dblp.org}
}

@inproceedings{DBLP:conf/cvpr/YinG0SDFP24,
  author = {Tianwei Yin and Micha{\"{e}}l Gharbi and Richard Zhang and Eli Shechtman and Fr{\'{e}}do Durand and William T. Freeman and Taesung Park},
  title = {One-Step Diffusion with Distribution Matching Distillation},
  booktitle = {Proceedings of the {IEEE/CVF} Conference on Computer Vision and Pattern Recognition},
  pages = {6613--6623},
  year = {2024},
  url = {https://doi.org/10.1109/CVPR52733.2024.00632},
  doi = {10.1109/CVPR52733.2024.00632},
  bibsource = {dblp computer science bibliography, https://dblp.org}
}

@inproceedings{DBLP:conf/nips/YinGP0SDF24,
  author = {Tianwei Yin and Micha{\"{e}}l Gharbi and Taesung Park and Richard Zhang and Eli Shechtman and Fr{\'{e}}do Durand and Bill Freeman},
  title = {Improved Distribution Matching Distillation for Fast Image Synthesis},
  booktitle = {Advances in Neural Information Processing Systems},
  volume = {37},
  year = {2024},
  url = {https://papers.nips.cc/paper_files/paper/2024/hash/54dcf25318f9de5a7a01f0a4125c541e-Abstract-Conference.html},
  bibsource = {dblp computer science bibliography, https://dblp.org}
}
